\documentclass[pmlr,twocolumn,10pt]{jmlr} 

\usepackage{booktabs}
\usepackage{siunitx}

\usepackage[switch]{lineno}

\theorembodyfont{\upshape}
\theoremheaderfont{\scshape}
\theorempostheader{:}
\theoremsep{\newline}

\usepackage{amsmath,amssymb}
\usepackage{url}
\usepackage{comment}
\usepackage{booktabs}
\usepackage{multirow}
\usepackage{ragged2e}
\usepackage{multicol}
\usepackage{algorithm}
\usepackage{algorithmic}
\usepackage{bm}
\usepackage{rotating}

\usepackage[justification=centering]{caption}
\usepackage{subcaption}

\usepackage{array}
\newcommand{\PreserveBackslash}[1]{\let\temp=\\#1\let\\=\temp}
\newcolumntype{C}[1]{>{\PreserveBackslash\centering}p{#1}}
\newcolumntype{R}[1]{>{\PreserveBackslash\raggedleft}p{#1}}
\newcolumntype{L}[1]{>{\PreserveBackslash\raggedright}p{#1}}

\def\eqref#1{equation~\ref{#1}}

\def\1{\bm{1}}

\DeclareMathAlphabet{\mathsfit}{\encodingdefault}{\sfdefault}{m}{sl}
\SetMathAlphabet{\mathsfit}{bold}{\encodingdefault}{\sfdefault}{bx}{n}

\newcommand{\std}[1]{\tiny{$\pm$#1}}

\jmlrvolume{287}
\jmlryear{2025}
\jmlrsubmitted{}
\jmlrpublished{}
\jmlrworkshop{Conference on Health, Inference, and Learning (CHIL) 2025} 

\title[Multi-View Contrastive Learning for Robust Domain Adaptation in Medical Time Series Analysis]{Multi-View Contrastive Learning for Robust Domain Adaptation in Medical Time Series Analysis}

\author{%
\Name{YongKyung Oh} 
\Email{yongkyungoh@mednet.ucla.edu}\\
\addr Medical \& Imaging Informatics (MII), UCLA
\AND
\Name{Alex A. T. Bui}\thanks{Corresponding Author}
\Email{buia@mii.ucla.edu}\\
\addr Medical \& Imaging Informatics (MII), UCLA
}

\begin{document}

\maketitle

\begin{abstract}
    Adapting machine learning models to medical time series across different domains remains a challenge due to complex temporal dependencies and dynamic distribution shifts. Current approaches often focus on isolated feature representations, limiting their ability to fully capture the intricate temporal dynamics necessary for robust domain adaptation. In this work, we propose a novel framework leveraging multi-view contrastive learning to integrate temporal patterns, derivative-based dynamics, and frequency-domain features. Our method employs independent encoders and a hierarchical fusion mechanism to learn feature-invariant representations that are transferable across domains while preserving temporal coherence. Extensive experiments on diverse medical datasets, including electroencephalogram (EEG), electrocardiogram (ECG), and electromyography (EMG) demonstrate that our approach significantly outperforms state-of-the-art methods in transfer learning tasks. By advancing the robustness and generalizability of machine learning models, our framework offers a practical pathway for deploying reliable AI systems in diverse healthcare settings. 
\end{abstract}

\paragraph*{Data and Code Availability}
This study uses publicly available datasets in medical and healthcare domains, including \textsc{SleepEEG}~\citep{kemp2000analysis} and \textsc{ECG}~\citep{clifford2017af} for pre-training, and \textsc{Epilepsy}~\citep{andrzejak2001indications}, \textsc{FD}~\citep{lessmeier2016condition}, \textsc{Gesture}~\citep{liu2009uwave}, and \textsc{EMG}~\citep{goldberger2000physiobank} for fine-tuning. 
The datasets used in this study are publicly accessible via their respective repositories, with detailed documentation included in the supplementary material. 
Additionally, implementation details and code repository\footnote{\url{https://github.com/yongkyung-oh/Multi-View_Contrastive_Learning}} are provided to facilitate reproducibility.

\paragraph*{Institutional Review Board (IRB)}
This research uses publicly available datasets that do not include identifiable personal information and, as such, does not require IRB approval. 

\section{Introduction}
Time series data form the backbone of modern healthcare monitoring and clinical decision-making, where the analysis of temporal patterns from diverse medical signals directly influences patient care outcomes and treatment strategies.
However, effective analysis of time series data poses significant challenges due to complex temporal dependencies, non-stationarity, irregularity, and distribution shifts that occur across different domains \citep{hyndman2018forecasting, hamilton2020time, oh2024stable}. These challenges are further amplified when models trained in one domain are applied to another, leading to degraded performance—a phenomenon known as \emph{domain shift}~\citep{li2017deeper, quinonero2022dataset}.

\emph{Domain adaptation} aims to address this issue by transferring knowledge learned from a large source domain to a small target domain, thereby enhancing model performance on the target data \citep{pan2009survey, farahani2021brief}. While domain adaptation has seen considerable success in fields like computer vision and natural language processing \citep{ganin2015unsupervised, tzeng2017adversarial}, its application to time series data remains less explored due to the inherent complexities of temporal information, including the alignment of temporal structures and the mitigation of shifts in both feature and label distributions \citep{wilson2020multi, shi2022deep}.

Traditional domain adaptation techniques often fall short in addressing the intricate temporal dependencies and dynamic distribution shifts characteristic of time series data, frequently resulting in suboptimal performance. Approaches that focus exclusively on minimizing statistical distances between source and target domains, such as Maximum Mean Discrepancy (MMD)-based methods, tend to neglect the temporal relationships essential for effective modeling and knowledge transfer \citep{long2015learning, cai2021time, liu2021adversarial}. To address these challenges, recent advancements in contrastive learning and self-supervised frameworks have shown promise by capturing invariant features through representations based on temporal and frequency information \citep{eldele2021time, ozyurt2023contrastive, zhang2024self}. 
Specifically, techniques such as feature alignment and adversarially-learned embeddings have demonstrated improved robustness against domain shifts, enhancing the applicability of these models to real-world scenarios \citep{jin2022domain, ragab2023adatime, oh2023sensor}.

Despite these advancements, existing methods predominantly focus on a single type of feature representation or decomposition, limiting their ability to fully capture the multifaceted temporal characteristics of time series data~\citep{zerveas2021transformer, cai2021time, yue2022ts2vec, lee2024soft}. Furthermore, many approaches inadequately address the need for invariant representations that simultaneously preserve temporal coherence, transfer knowledge, and adapt to dynamic distribution shifts \citep{sun2016deep, weber2021transfer, oh2024multichannel}.

To further ground our research in practical and impactful applications, our study focuses on the medical domain, where time series data play a pivotal role. Medical time series, such as electrocardiograms (ECGs), electroencephalograms (EEGs), electronic health records (EHRs), and patient monitoring data often exhibit complex temporal dependencies and significant distribution shifts due to variations in patient populations, equipment, and clinical settings \citep{shickel2017deep, purushotham2018benchmarking, harutyunyan2019multitask, xie2022deep}. These shifts pose barriers to deploying machine learning models across diverse healthcare environments. 
Existing domain adaptation techniques are particularly limited in this context due to the need for preserving temporal coherence while addressing  sensitive and dynamic distribution shifts. 

Here, we propose a novel self-supervised framework for time series feature adaptation that leverages complementary representations learned through multi-view contrastive learning. Specifically, our approach integrates information from three distinct features: the \emph{temporal-feature}, which captures inherent patterns and trends within the raw data; the \emph{derivative-feature}, which characterizes local dynamics and trend changes over time; and the \emph{frequency-feature}, which provides insights into global spectral features of the time series. By combining these complementary features, our framework aims to learn \emph{domain-invariant representations} that are robust to distribution shifts while preserving temporal coherence.
Extensive experiments on benchmark datasets demonstrate the superiority of our approach over state-of-the-art methods, validating the advantages of integrating complementary representations and leveraging contrastive learning.

The remainder of the paper is organized as follows. Section~\ref{sec:Related_works} reviews related work in time series domain adaptation and contrastive learning approaches. Section~\ref{sec:Methodology} introduces our proposed framework, detailing the multi-view feature extraction and hierarchical fusion mechanisms. Section~\ref{sec:Experiment} presents comprehensive experimental evaluations on medical time series datasets including `one-to-one scenario' and `one-to-many scenario'. 
Section~\ref{sec:Discussion} examines practical implications and limitations, while Section~\ref{sec:Conclusion} outlines conclusions and future research directions.

\begin{figure*}[htbp]
\floatconts
  {fig:overview} 
  {\caption{Overview of the proposed framework. The dotted box represents the multi-view contrastive pre-training phase using \(\mathcal{D}_S\), while the dashed box corresponds to the fine-tuning phase using \(\mathcal{D}_T\).}} 
  {%
    \centering
    \includegraphics[scale=0.47]{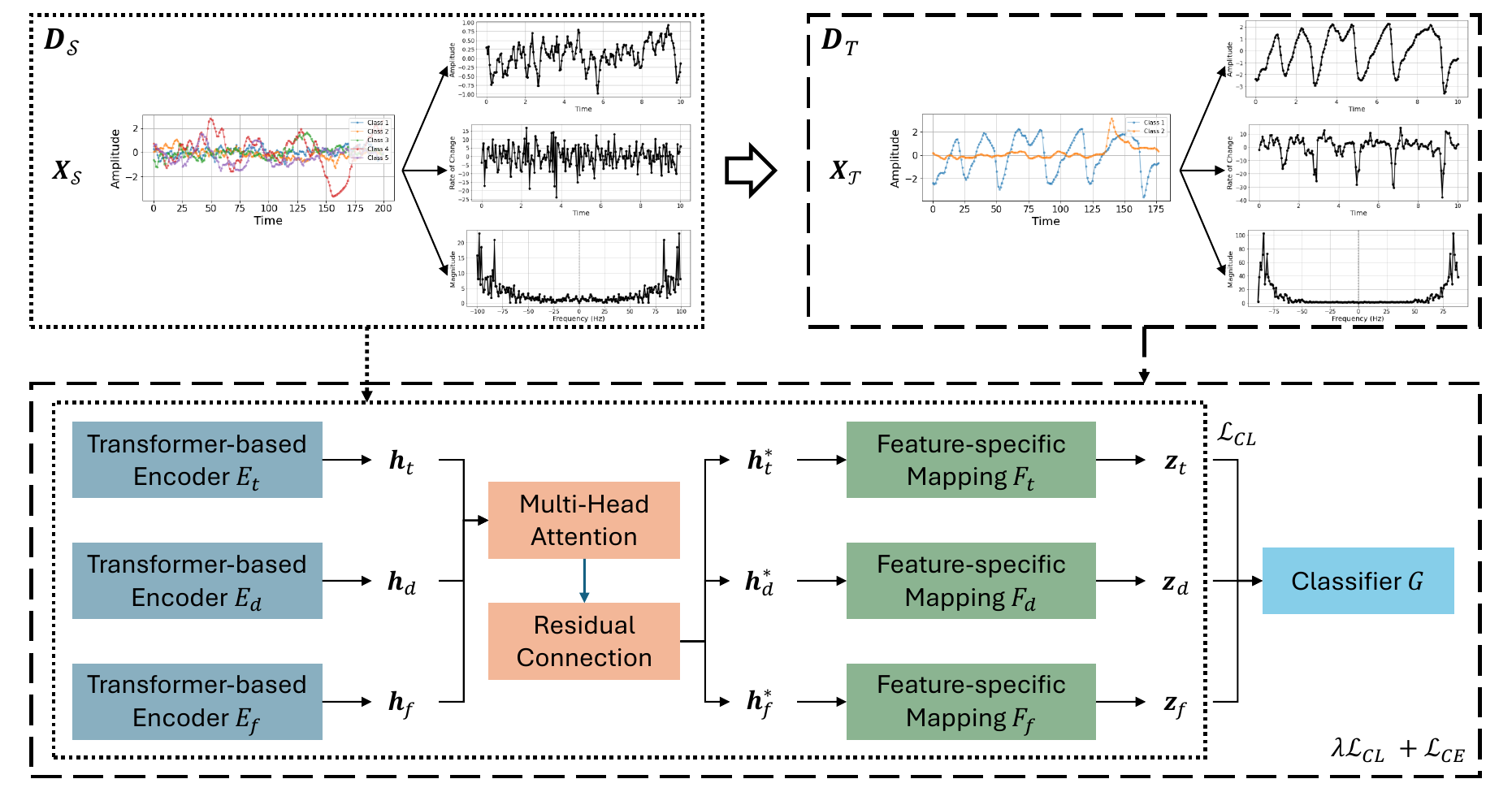} 
    \vspace{-10pt}
  }
\end{figure*}

\section{Related works}\label{sec:Related_works}
Time series domain adaptation has been extensively studied as a means to address distribution shifts between source and target domains \citep{ben2010theory, purushotham2017variational, ragab2023adatime, zhang2024self}. 
Various recent approaches have been explored for time series domain adaptation, including sparse associative structure alignment for inter-variable relationships, contrastive learning for temporal and contextual patterns, and consistency enforcement between time and frequency domains \citep{cai2021time,eldele2021time,zhang2022self}. 
These methods highlight the importance of leveraging multiple feature representations to effectively address domain shifts in time series data.

Contrastive learning has emerged as a powerful tool for self-supervised representation learning, showing impressive results across various domains \citep{chen2020simple, he2020momentum, tian2020contrastive}. In time series analysis, contrastive methods have been employed to learn robust representations by defining positive and negative pairs based on temporal proximity or data augmentations \citep{woo2022cost, zhang2022self}. These approaches highlight the potential of leveraging invariances within temporal data to improve generalization across domains.

In the medical domain, domain adaptation for time series faces unique challenges. Medical time series datasets are characterized by high inter-patient variability, non-stationarity due to physiological conditions, and differences in data collection settings \citep{he2023domain, wang2024contrast}. These factors exacerbate distribution shifts and complicate the adaptation process. Approaches like domain adversarial networks and hierarchical feature alignment have been applied to mitigate these challenges, but often fail to fully address the critical need for preserving medical context and interpretability. 

Still, existing methods often focus on a single type of feature or decomposition, limiting their ability to fully capture the inherent temporal characteristics of time series data. They may overlook the benefits of integrating complementary representations from multiple domains—such as time, derivative, and frequency domains—which can provide a more comprehensive understanding of the data. 
By focusing on these limitations, our study contributes to bridging the gap between robust representation learning and effective domain adaptation for time series in both general and medical contexts.

\section{Methodology}\label{sec:Methodology}

\subsection{Problem Formulation}
Let $\mathcal{D}_{\mathcal{S}} = \{ (\bm{X}_{\mathcal{S}}^{(i)}, y_{\mathcal{S}}^{(i)}) \}_{i=1}^{N_{\mathcal{S}}}$ denote the source domain dataset, where $\bm{X}_{\mathcal{S}}^{(i)} \in \mathbb{R}^{T_{\mathcal{S}} \times d_{\mathcal{S}}}$ is a multivariate time series sample with $T_{\mathcal{S}}$ timesteps and $d_{\mathcal{S}}$ variables, and $y_{\mathcal{S}}^{(i)} \in \mathcal{Y}_{\mathcal{S}}$ is the corresponding label. Similarly, let $\mathcal{D}_{\mathcal{T}} = \{ \bm{X}_{\mathcal{T}}^{(i)}, y_{\mathcal{T}}^{(i)} \}_{i=1}^{N_{\mathcal{T}}}$ denote the target domain dataset, where $\bm{X}_{\mathcal{T}}^{(i)} \in \mathbb{R}^{T_{\mathcal{T}} \times d_{\mathcal{T}}}$ and $y_{\mathcal{T}}^{(i)} \in \mathcal{Y}_{\mathcal{T}}$. Our goal is to learn encoders $E(\cdot)$, feature extractor $F(\cdot)$ and a classifier $G(\cdot)$ such that the classifier $G(F(E(\bm{X}_{\mathcal{T}})))$ performs well on the target domain, despite the distribution shift between $\mathcal{D}_{\mathcal{S}}$ and $\mathcal{D}_{\mathcal{T}}$.

Note that for contrastive pre-training, we treat the source domain \(\mathcal{D}_{\mathcal{S}}\) as unlabeled, and thus the labels \(y_{\mathcal{S}}^{(i)}\) are not utilized during representation learning.

\subsection{Overview of the Proposed Framework}
As shown in \figureref{fig:overview}, our framework consists of two phases: multi-view contrastive pre-training and fine-tuning. In the pre-training phase, we leverage three complementary views of time series data from source domain $\mathcal{D}_S$: temporal, derivative, and frequency domains. Each view is processed through its respective encoder ($E_t$, $E_d$, $E_f$) and projector ($F_t$, $F_d$, $F_f$) to obtain view-specific representations in the contrastive learning space. These representations are optimized using contrastive loss $\mathcal{L}_{CL}$ to capture the inherent relationships across different views. During the fine-tuning phase, we transfer the pre-trained encoders and projectors to the target domain $\mathcal{D}_T$ and introduce a classification head $G$ that maps the concatenated representations to the target task output, optimized using {Cross-entropy} loss $\mathcal{L}_{CE}$. The detailed components and methodology of our framework are elaborated in the following sections.

\subsection{Multi-view Feature Extraction}\label{sec:feature}

\subsubsection{Temporal-feature Encoder}
The temporal-feature encoder $E_t(\cdot)$ captures inherent patterns and trends within the raw time series data. The temporal-feature representation is obtained as:
\begin{align*}
    \bm{h}_t = E_t(\bm{X}) \in \mathbb{R}^{L \times D},
\end{align*}
where \(L\) represents the length of time series and can be \(T_\mathcal{S}\) or \(T_\mathcal{T}\),  and \(D\) is the hidden dimension.

\subsubsection{Derivative-feature Encoder}
To capture local dynamics and shifts in trends over time, we compute derivative features using finite difference interpolation. 
Given the time series $\bm{X}$, we approximate the first-order derivative $\dot{\bm{X}}$ as follows:
\begin{align*}
    \dot{\bm{X}}_t = \frac{3\bm{X}_t - 4\bm{X}_{t-1} + \bm{X}_{t-2}}{2\Delta t},
\end{align*}
where $\Delta t$ is the time interval between consecutive samples, and $t = 2, \ldots, T$.
An interpolation technique is used for derivative computation due to its smoothness properties and robustness to noise, as discussed in \citet{morrill2022choice}.
The derivative-feature representation is then obtained by passing $\dot{\bm{X}}$ through the encoder $E_d(\cdot)$ and padding:
\begin{align*}
    \bm{h}_d = E_d(\dot{\bm{X}}) \in \mathbb{R}^{L \times D}.
\end{align*}

\subsubsection{Frequency-feature Encoder}
The frequency-feature encoder $E_f(\cdot)$ captures global spectral characteristics of the time series. We compute the frequency-feature representation using the Fast Fourier Transform (FFT). We take the magnitude of the FFT coefficients to obtain the amplitude spectrum:
\begin{align*}
    \bm{X}_{\text{freq}} = |\text{FFT}(\bm{X})|.
\end{align*}
The frequency-feature representation is then extracted using the encoder $E_f(\cdot)$:
\begin{align*}
    \bm{h}_f = E_f(\bm{X}_{\text{freq}}) \in \mathbb{R}^{L \times D}.
\end{align*}

We implement $E_t$, $E_d$, and $E_f$, using a Transformer encoder architecture \citep{vaswani2017attention}, which effectively models long-range dependencies through self-attention mechanisms. By utilizing these encoders, our framework extracts rich features from each feature, which are later fused to form a comprehensive representation of the time series data.

\subsection{Hierarchical Feature Fusion}

To integrate the representations from the three features—time, derivative, and frequency—we employ a hierarchical fusion mechanism designed to capture complex interactions among these features. 
Although these features are derived from the same input data, the fusion process can enhance mutual information by integrating complementary representations~\citep{oh2024deep,qiu2024multimodal}.
This fusion operates by applying multi-head attention across the stacked representations, allowing the model to focus on salient features and interactions between different features, discussed in Section~\ref{sec:feature}.

Given the multi-view feature representations using encoders, \(\bm{h}_t, \bm{h}_d, \bm{h}_f \in \mathbb{R}^{L \times D}\), where \(L\) is the sequence length, and \(D\) is the hidden dimension, we stack the three representations along a new dimension:
\begin{align*}
    \bm{H} = \text{stack}(\bm{h}_t, \bm{h}_d, \bm{h}_f) \in \mathbb{R}^{L \times 3 \times D}.
\end{align*}
We then apply multi-head attention (MHA) on the stacked feature \(\bm{H}\),
so that the attention is computed across the feature dimension. 
After the attention mechanism, a residual connection and subsequent layer normalization are applied:
\begin{align*}
    \bm{H}_{\text{output}} = \text{LayerNorm}(\bm{H} + \text{MHA}(\bm{H})).
\end{align*}
We extract the updated representations for each feature by splitting \(\bm{H}_{\text{output}}\) along the feature dimension:
\begin{align*}
    \bm{H}_{\text{output}} = [\bm{h}_t^{*}, \bm{h}_d^{*}, \bm{h}_f^{*}],
\end{align*}
where \(\bm{h}_t^{*}, \bm{h}_d^{*}, \bm{h}_f^{*} \in \mathbb{R}^{L \times D}\) are the updated feature representations. These representations now contain information not only from their respective features but also from interactions with the other features.

To obtain the final representation for downstream task, we process each updated feature representation through a feature-specific mapping \(F_{k}(\cdot)\), which are learnable feed-forward networks and averaging layer: 
\begin{align*}
    \bm{z}_k = F_{k}(\bm{h}_k^{*}) \quad \text{for} \quad k \in \{t, d, f\}.
\end{align*}
By applying attention across stacked representations, the model learns to capture dependencies and interactions between different features. Additionally, the attention mechanism allows the model to focus on informative features between each other, improving the quality of the fused representation.

\subsection{Optimization with Combined Losses}
Our proposed framework leverages two key loss functions—feature-specific contrastive losses and supervised classification loss—optimized in two sequential stages: pre-training on the source dataset and fine-tuning on the target dataset.

\paragraph{Feature-Specific Contrastive Losses}
To enhance the quality of feature representations, we employ a contrastive loss for each feature type (temporal, derivative, and frequency). Positive and negative pairs are constructed using time series-specific augmentations as suggested by \citet{zhang2022self}. For each feature $k \in \{t, d, f\}$, let $\bm{z}_k$ and $\tilde{\bm{z}}_k$ represent the original and augmented feature representations, respectively. The contrastive InfoNCE is defined as:
\begin{align*}
    \mathcal{L}_{\text{CL}}^k = - \frac{1}{N} \sum_{i=1}^{N} \log \frac{\exp(\text{sim}(\bm{z}_k^{(i)}, \tilde{\bm{z}}_k^{(i)})/\tau)}{\sum_{j=1}^{N} \exp(\text{sim}(\bm{z}_k^{(i)}, \tilde{\bm{z}}_k^{(j)})/\tau)},
\end{align*}
where $\text{sim}(\bm{u}, \bm{v}) = \frac{\bm{u}^\top \bm{v}}{\|\bm{u}\| \|\bm{v}\|}$ is the cosine similarity, $\tau$ is the temperature parameter set to 0.07, as suggested by \citet{he2020momentum}, and $N$ is the batch size. The total contrastive loss is the sum of feature-specific losses:
\begin{align*}
    \mathcal{L}_{\text{CL}} = \sum_{k \in \{t, d, f\}} \mathcal{L}_{\text{CL}}^k.
\end{align*}

\paragraph{Supervised Classification Loss}
For labeled target data, we use the supervised cross-entropy loss to train the classifier. Let $\bm{z}_{\text{combined}} = [\bm{z}_t, \bm{z}_d, \bm{z}_f]$ be the concatenated feature representations. For the classifier $G(\cdot)$, the predicted label for the $i$-th sample is:
\begin{align*}
    \hat{y}^{(i)} = G(\bm{z}_{\text{combined}}^{(i)}).
\end{align*}
The supervised classification loss is computed as:
\begin{align*}
    \mathcal{L}_{\text{CE}} = -\frac{1}{N} \sum_{i=1}^{N} \sum_{c=1}^{C} \mathbb{I}_{[y^{(i)} = c]} \log p^{(i)}_{c},
\end{align*}
where $C$ is the number of classes, $\mathbb{I}_{[\cdot]}$ is the indicator function, and $p^{(i)}_{c}$ denotes the predicted probability for class $c$.

\paragraph{Optimization Procedure}
The training process involves two stages:
\begin{enumerate}
\item \textbf{Pre-training on Source Dataset}: The model is trained on the source dataset using only the InfoNCE loss to learn feature-invariant representations across $\{t, d, f\}$. This stage optimizes the encoders $E_k$ and projection heads $F_k$ for each feature type $k \in \{t, d, f\}$, ensuring that each feature view preserves its intra-view information independently through contrastive learning.

\item \textbf{Fine-tuning on Target Dataset}: The pre-trained model is fine-tuned on the target dataset using a weighted combination of the InfoNCE loss and the cross-entropy loss. This stage trains the encoders $E_k$, projection heads $F_k$, and the classifier $G$:
\begin{align*}
    \mathcal{L}_{\text{total}} = \frac{1}{N} \sum_{i=1}^{N} \left(\lambda \mathcal{L}_{\text{CL}}^{(i)} + \mathcal{L}_{\text{CE}}^{(i)} \right),
\end{align*}
where $\lambda$ is a hyperparameter balancing the two losses, set to 0.1. This fine-tuning phase leverages hierarchical fusion to integrate multi-view information, capturing inter-view dependencies that enhance predictive performance.
\end{enumerate}

The losses are optimized using the Adam optimizer \citep{kingma2014adam}, ensuring efficient convergence during both stages. The rationale behind our architectural design and its theoretical foundation are provided in the Appendix~\ref{appendix:theoretical}.

\begin{table*}[!htb]
\small\centering\captionsetup{justification=centering, skip=5pt}
\caption{Description of dataset statistics. For the number of samples in the target (fine-tuning) dataset, ``$n_1$\;/\;$n_2$\;/\;$n_3$" indicates $n_1$ samples for fine-tuning, $n_2$ samples for validation, and $n_3$ samples for testing. 
}
\label{tab:data_statistics}
\begin{tabular}{lcccccc}
\toprule
\multicolumn{1}{c}{\textbf{Dataset}} & \multicolumn{1}{l}{\# Samples} & \multicolumn{1}{l}{\# Channels} & \multicolumn{1}{l}{\# Classes} & \multicolumn{1}{l}{Length} & \multicolumn{1}{l}{Freq (Hz)} \\
\midrule
\textsc{SleepEEG}  & 371,055 & 1 & 5 & 200 & 100 \\
\textsc{ECG}  & 43,673 & 1 & 4 & 1,500 & 300 \\
\midrule
\textsc{Epilepsy}  & 60\;/\;20\;/\;11,420 & 1 & 2 & 178 & 174 \\
\textsc{FD} & 60\;/\;21\;/\;13,559 & 1 & 3 & 5,120 & 64K \\
\textsc{Gesture} & 320\;/\;120\;/\;120 & 3 & 8 & 315 & 100 \\
\textsc{EMG} & 122\;/\;41\;/\;41 & 1 & 3 & 1,500 & 4,000 \\
\bottomrule
\end{tabular}
\end{table*}
\begin{figure*}[htb]
\floatconts
  {fig:sample}
  {\caption{Representative samples from each dataset, with different colors indicating distinct classes}}
  {%
    \subfigure[\textsc{SleepEEG}]{\label{fig:sleepEEG}%
      \includegraphics[width=0.32\linewidth]{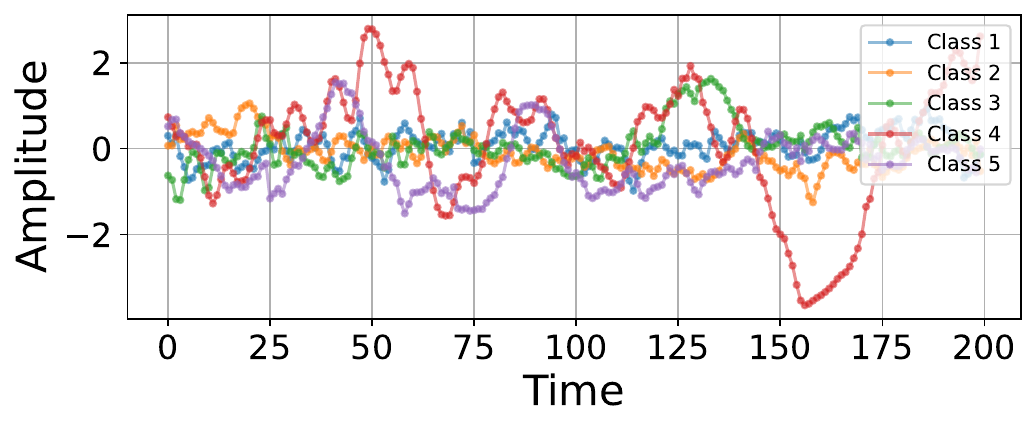}}%
    \hfil
    \subfigure[\textsc{ECG}]{\label{fig:ECG}%
      \includegraphics[width=0.32\linewidth]{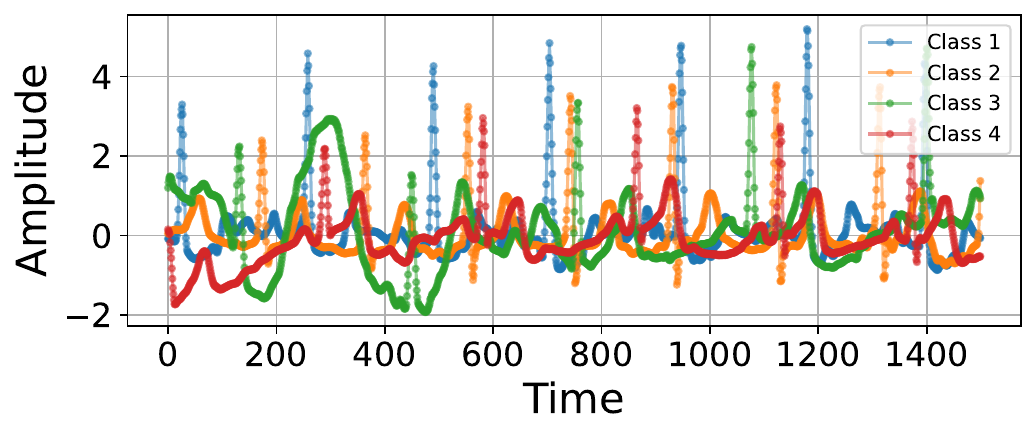}}%
    \hfil
    \subfigure[\textsc{Epilepsy}]{\label{fig:Epilepsy}%
      \includegraphics[width=0.32\linewidth]{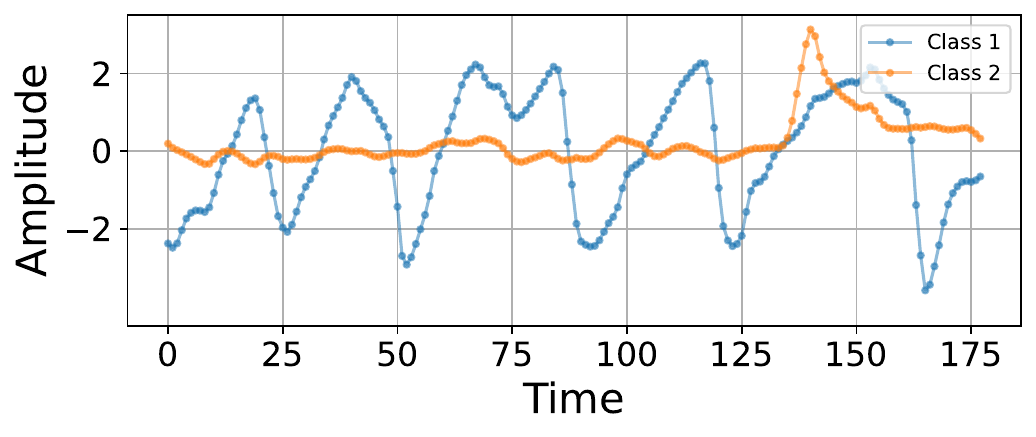}}%
    \\ \medskip
    \subfigure[\textsc{FD}]{\label{fig:FD}%
      \includegraphics[width=0.32\linewidth]{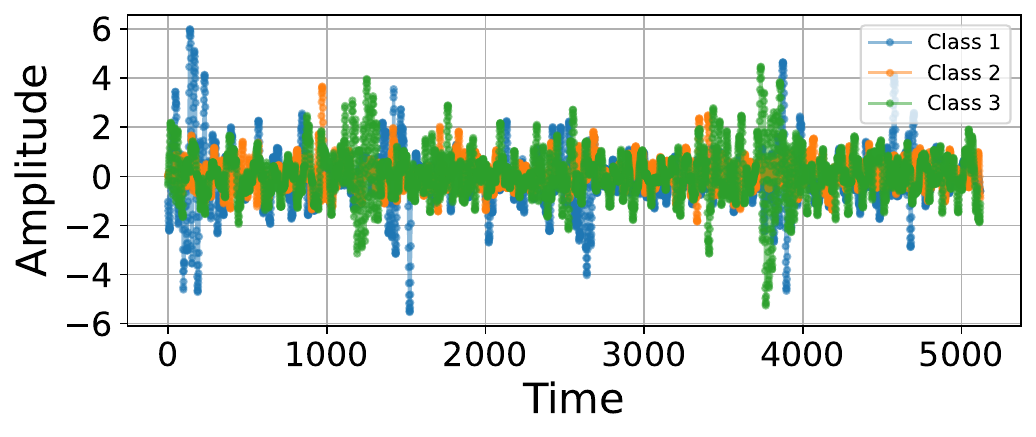}}%
    \hfil
    \subfigure[\textsc{Gesture}]{\label{fig:Gesture}%
      \includegraphics[width=0.32\linewidth]{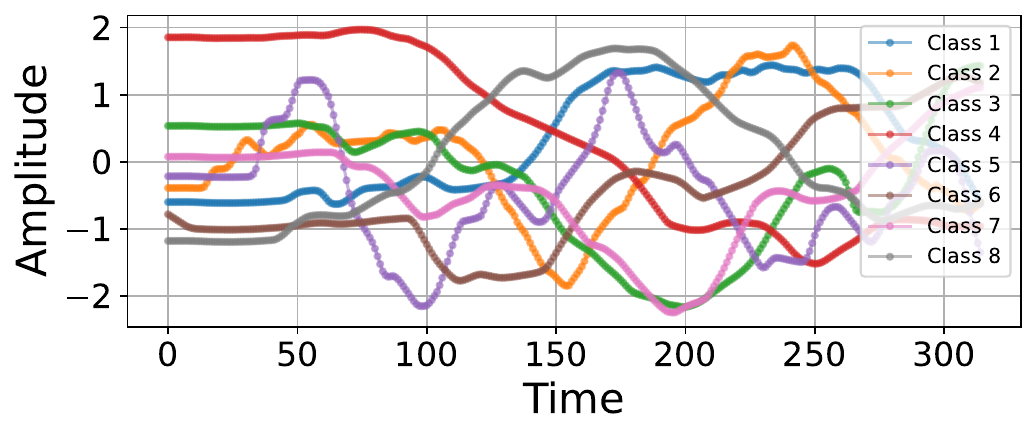}}%
    \hfil
    \subfigure[\textsc{EMG}]{\label{fig:EMG}%
      \includegraphics[width=0.32\linewidth]{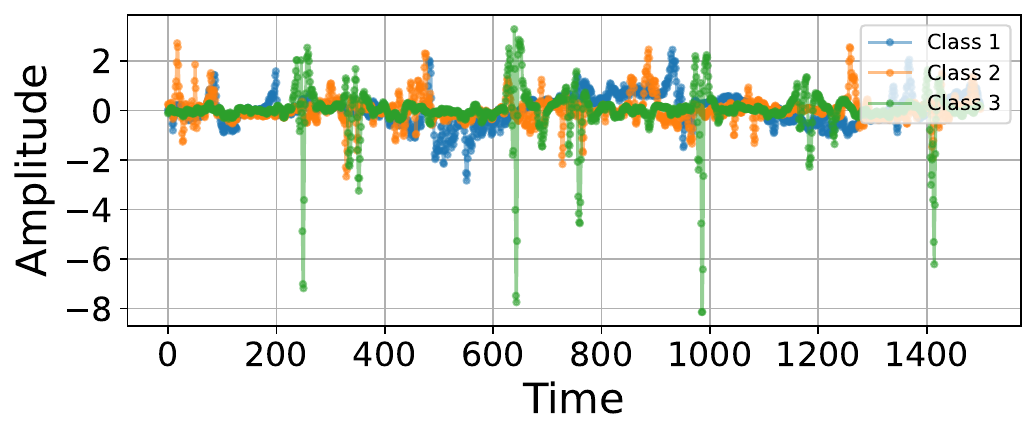}}%
    \vspace{-10pt}
  }
\end{figure*}

\section{Experiment}\label{sec:Experiment}
\subsection{Experimental Setting}
We evaluate our framework following the experimental pipeline introduced in \citet{zhang2022self} and \citet{dong2024simmtm}. Preprocessed data can be accessed through corresponding repository\footnote{\url{https://github.com/mims-harvard/TFC-pretraining}}.
We utilize \textsc{SleepEEG} as the source domain for pre-training, chosen for its rich temporal patterns and substantial data volume. The pre-trained model is then fine-tuned separately on four target domains: \textsc{Epilepsy}, \textsc{FD}, \textsc{Gesture}, and \textsc{EMG}. Additional to that, we use \textsc{ECG} as source dataset for sensitivity analysis.

Our experimental setup allows us to assess the model's transferability across diverse domains characterized by varying sampling rates, channel dimensions, sequence durations, and class compositions. 

Data Statistics of each dataset are described in Table~\ref{tab:data_statistics}. Representative samples from each dataset are illustrated in \figureref{fig:sample}, with comprehensive dataset descriptions provided in the Appendix~\ref{appendix:experimental}.

\subsection{Benchmark Methods}

Various approaches have been developed for time series representation learning, each with distinct characteristics.
\textbf{TST}~\citep{zerveas2021transformer} uses a Transformer-based framework for multivariate time series. 
\textbf{TS-SD}~\citep{shi2021self} applies self-supervised learning with sliding windows and dynamic time warping. 
\textbf{SimCLR}~\citep{tang2020exploring} adapts a prominent contrastive learning framework~\citep{chen2020simple} from computer vision to time series. 
\textbf{TS-TCC}~\citep{eldele2021time} leverages temporal and contextual contrasting. 
\textbf{CLOCS}~\citep{kiyasseh2021clocs} focuses on contrastive learning for cardiac signals in the medical domain. 
\textbf{Mixing-up}~\citep{wickstrom2022mixing} utilizes time series mixing for data augmentation. 
\textbf{TS2Vec}~\citep{yue2022ts2vec} provides a universal framework using hierarchical contrastive loss. 
\textbf{TF-C}~\citep{zhang2022self} emphasizes temporal-frequency contrasting for self-supervised learning. 
\textbf{CoST}~\citep{woo2022cost} disentangles seasonal-trend representations through contrastive learning. 
\textbf{Ti-MAE}~\citep{li2023ti} employs a masked autoencoder for self-supervised learning. 
\textbf{SimMTM}~\citep{dong2024simmtm} uses masked time points with weighted neighbor aggregation for representation learning.

For the fair performance comparisons, we used identical pipeline and reported performance metrics from \citet{zhang2022self} and \citet{dong2024simmtm}.

\begin{table*}[!htb]
\small\centering\captionsetup{justification=centering, skip=5pt}
\caption{Performance comparison with benchmark method: \textsc{SleepEEG} $\rightarrow$ \textsc{Epilepsy}\\ 
(mean ± standard deviation over five runs; $0.000$ indicates unreported variance in the referenced studies)
}\label{tab:result1_full}
\begin{tabular}{@{}lC{2.1cm}C{2.1cm}C{2.1cm}C{2.1cm}@{}}
\toprule
\multicolumn{1}{c}{\textbf{Method\hspace{1.0em}\textbackslash\hspace{1.0em}Metric}} & \textbf{Accuracy} & \textbf{Precision} & \textbf{Recall}  & \textbf{F1 score} \\ \midrule
TST~\citep{zerveas2021transformer}                                                        & 0.802\std{0.000}  & 0.401\std{0.000}   & 0.500\std{0.000} & 0.445\std{0.000}  \\
TS-SD~\citep{shi2021self}                                                      & 0.895\std{0.052}  & 0.802\std{0.224}   & 0.765\std{0.149} & 0.777\std{0.186}  \\
SimCLR~\citep{tang2020exploring}                                                     & 0.907\std{0.034}  & 0.922\std{0.017}   & 0.786\std{0.107} & 0.818\std{0.100}  \\
TS-TCC~\citep{eldele2021time}                                                     & 0.925\std{0.010}  & 0.945\std{0.005}   & 0.818\std{0.026} & 0.863\std{0.022}  \\
CLOCS~\citep{kiyasseh2021clocs}                                                      & 0.951\std{0.003}  & 0.930\std{0.007}   & 0.913\std{0.017} & 0.921\std{0.007}  \\
Mixing-up~\citep{wickstrom2022mixing}                                                  & 0.802\std{0.000}  & 0.401\std{0.000}   & 0.500\std{0.000} & 0.445\std{0.000}  \\
TS2Vec~\citep{yue2022ts2vec}                                                     & 0.940\std{0.004}  & 0.906\std{0.012}   & 0.904\std{0.012} & 0.905\std{0.007}  \\
TF-C~\citep{zhang2022self}                                                       & 0.950\std{0.025}  & \textbf{0.946\std{0.011}}   & 0.891\std{0.022} & 0.915\std{0.053}  \\
CoST~\citep{woo2022cost}                                                       & 0.884\std{0.000}  & 0.882\std{0.000}   & 0.723\std{0.000} & 0.769\std{0.000}  \\
Ti-MAE~\citep{li2023ti}                                                     & 0.897\std{0.000}  & 0.724\std{0.000}   & 0.675\std{0.000} & 0.686\std{0.000}  \\
SimMTM~\citep{dong2024simmtm}                                                     & 0.955\std{0.000}  & 0.934\std{0.000}   & 0.923\std{0.000} & 0.928\std{0.000}  \\ \midrule
\textbf{Proposed method}                                    & \textbf{0.956\std{0.002}}  & 0.936\std{0.004}   & \textbf{0.935\std{0.004}} & \textbf{0.931\std{0.003}}  \\ \bottomrule
\end{tabular}
%
\\ \bigskip\bigskip
%
\small\centering\captionsetup{justification=centering, skip=5pt}
\caption{Sensitivity analysis on \textsc{Epilepsy} with a variety of learning strategies: different source datasets (\textsc{SleepEEG} and \textsc{ECG}), self-supervised learning, and random initialization (without pre-training) \\ 
(For each experimental scenario, classifier $G(\cdot)$ was initialized and subsequently trained by target dataset. Default setting fine-tunes of all $E_k(\cdot)$, $F_k(\cdot)$, and $G(\cdot)$.
`freeze scenario' indicates that $E_k(\cdot)$ and $F_k(\cdot)$ are fixed after pre-training, while only $G(\cdot)$ are optimized during fine-tuning, where $k \in \{t,d,f\}$.)
}\label{tab:result1_sensitivity}
\begin{tabular}{@{}lC{2.1cm}C{2.1cm}C{2.1cm}C{2.1cm}@{}}
\toprule
\multicolumn{1}{c}{\textbf{Learning Strategy\hspace{1.0em}\textbackslash\hspace{1.0em}Metric}} & \textbf{Accuracy} & \textbf{Precision} & \textbf{Recall}  & \textbf{F1 score} \\ \midrule
Source: \textsc{SleepEEG}                                   & 0.956\std{0.002}  & 0.936\std{0.004}   & 0.935\std{0.004} & 0.931\std{0.003}  \\
-- freeze $E_k(\cdot)$ and $F_k(\cdot)$                                        & 0.952\std{0.005}  & 0.925\std{0.012}   & 0.931\std{0.004} & 0.924\std{0.006}  \\ \midrule
Source: \textsc{ECG}                                        & 0.953\std{0.002}  & 0.934\std{0.004}   & 0.940\std{0.008} & 0.927\std{0.003}  \\
-- freeze $E_k(\cdot)$ and $F_k(\cdot)$                                        & 0.952\std{0.002}  & 0.932\std{0.012}   & 0.935\std{0.006} & 0.924\std{0.002}  \\ \midrule
Source: \textsc{Epilepsy}                                   & 0.953\std{0.002}  & 0.931\std{0.009}   & 0.941\std{0.003} & 0.926\std{0.002}  \\
-- freeze $E_k(\cdot)$ and $F_k(\cdot)$                                        & 0.953\std{0.003}  & 0.928\std{0.012}   & 0.939\std{0.004} & 0.925\std{0.002}  \\ \midrule
Random initialization                                                & 0.948\std{0.004}  & 0.911\std{0.014}   & 0.936\std{0.006} & 0.920\std{0.004}  \\ \bottomrule
\end{tabular}
%
\\ \bigskip\bigskip
%
\small\centering\captionsetup{justification=centering, skip=5pt}
\caption{Comprehensive analysis of the proposed framework's components on \textsc{Epilepsy} \\ 
(1) hierarchical feature fusion, (2) fine-tuning loss configurations with $\lambda$, and (3) combinations of features $\left\{\bm{h}_t, \bm{h}_d, \bm{h}_f \right\}$. (The proposed method utilizes all three features with $\lambda=0.1$ as the default configuration.)
}\label{tab:result1_ablation}
\begin{tabular}{@{}lC{2.1cm}C{2.1cm}C{2.1cm}C{2.1cm}@{}}
\toprule
\multicolumn{1}{c}{\textbf{Model Component\hspace{1.0em}\textbackslash\hspace{1.0em}Metric}} & \textbf{Accuracy} & \textbf{Precision} & \textbf{Recall}  & \textbf{F1 score} \\ \midrule
Proposed method                                             & 0.956\std{0.002}  & 0.936\std{0.004}   & 0.935\std{0.004} & 0.931\std{0.003}  \\
-- use   $\mathcal{L}_{CE}$ only                            & 0.955\std{0.003}  & 0.930\std{0.005}   & 0.935\std{0.003} & 0.929\std{0.004}  \\ \midrule
w/o feature   fusion                                        & 0.956\std{0.002}  & 0.934\std{0.007}   & 0.936\std{0.003} & 0.930\std{0.004}  \\
-- use   $\mathcal{L}_{CE}$ only                            & 0.952\std{0.006}  & 0.922\std{0.015}   & 0.936\std{0.003} & 0.925\std{0.007}  \\ \midrule
$\left\{\bm{h}_t,   \bm{h}_d\right\}$                       & 0.945\std{0.003}  & 0.925\std{0.011}   & 0.916\std{0.006} & 0.911\std{0.004}  \\
$\left\{\bm{h}_t,   \bm{h}_f\right\}$                       & 0.952\std{0.003}  & 0.926\std{0.012}   & 0.928\std{0.010} & 0.925\std{0.003}  \\
$\left\{\bm{h}_d,   \bm{h}_f\right\}$                       & 0.955\std{0.003}  & 0.934\std{0.006}   & 0.935\std{0.005} & 0.928\std{0.004}  \\ \midrule
$\bm{h}_t$   only                                           & 0.931\std{0.008}  & 0.890\std{0.021}   & 0.898\std{0.005} & 0.892\std{0.009}  \\
$\bm{h}_d$   only                                           & 0.940\std{0.002}  & 0.907\std{0.004}   & 0.910\std{0.004} & 0.905\std{0.002}  \\
$\bm{h}_f$   only                                           & 0.955\std{0.003}  & 0.929\std{0.009}   & 0.936\std{0.005} & 0.930\std{0.005}  \\ \bottomrule
\end{tabular}
\vspace{\fill}
\end{table*}

\subsection{Results with One-to-one Scenario}
In this scenario, the framework is pre-trained on the \textsc{SleepEEG} dataset and fine-tuned on the \textsc{Epilepsy} dataset, both comprising single-channel EEG signals. The \textsc{SleepEEG} dataset provides recordings focused on sleep stages, while the \textsc{Epilepsy} dataset involves seizure and non-seizure classification. Despite both datasets being single-channel EEG, they differ in key aspects, including the scalp electrode positions used for recording, the physiological phenomena tracked (sleep patterns versus epilepsy activity), and the patient populations involved in the data collection.

Table~\ref{tab:result1_full} demonstrates a detailed performance comparison of the proposed method against several benchmark approaches for domain adaptation, specifically transferring from \textsc{SleepEEG} to \textsc{Epilepsy}. The proposed method outperforms all benchmarks across accuracy, precision, recall, and F1 score, achieving an F1 score of 0.931 ± 0.003, indicating its robustness and superior ability for domain adaptation task in medical time series analysis. 

Table~\ref{tab:result1_sensitivity} explores the sensitivity of the proposed method on the \textsc{Epilepsy} dataset with various learning strategies. Pre-training on the \textsc{SleepEEG} dataset yields the highest F1 score (0.931 ± 0.003) when using fine-tuning with trainable encoders. Freezing the encoders results in a slight decrease in performance (F1: 0.924 ± 0.004), demonstrating the importance of continued encoder updates during fine-tuning.
Pre-training on the \textsc{ECG} dataset also achieves competitive results, 
indicating the general adaptability of the proposed method across pre-training sources. However, pre-training directly on \textsc{Epilepsy}, which is self-supervised manner, shows slightly lower performance, suggesting that leveraging a larger, more complex source dataset like \textsc{SleepEEG} provides better feature generalization for the target domain.
Interestingly, even with random initialization, the proposed method achieves an F1 score of 0.920 ± 0.004, indicating the robustness of its hierarchical feature fusion mechanism. This robustness highlights the method’s ability to perform well even in the absence of domain-specific pre-training, though optimal results are achieved with targeted pre-training strategies.

Table~\ref{tab:result1_ablation} examines the contributions of different components in the proposed framework on the \textsc{Epilepsy} dataset, focusing on hierarchical feature fusion, fine-tuning loss configurations, and feature view combinations. The full proposed method, utilizing all features ($\bm{h}_t$, $\bm{h}_d$, $\bm{h}_f$) with hierarchical fusion and fine-tuning loss ($\mathcal{L}_{\text{CE}}$), achieves the highest F1 score. 
The results across these settings collectively demonstrate the robustness, adaptability, and effectiveness of the proposed multi-view contrastive learning framework in addressing domain adaptation challenges in time series analysis.

\begin{table*}[!htb]
\small\centering\captionsetup{justification=centering, skip=5pt}
\caption{F1 score comparison between our method and benchmark approaches. All models are pre-trained on \textsc{SleepEEG} dataset and evaluated through fine-tuning on four different target datasets. \\ 
(mean ± standard deviation over five runs; $0.000$ indicates unreported variance in the referenced studies)
}
\label{tab:result1}
\begin{tabular}{@{}lC{2.1cm}C{2.1cm}C{2.1cm}C{2.1cm}@{}}
\toprule
\multicolumn{1}{c}{\textbf{Method\hspace{1.0em}\textbackslash\hspace{1.0em}Target Dataset}} & \textbf{Epilepsy} & \textbf{FD}      & \textbf{Gesture} & \textbf{EMG}     \\ \midrule
TST~\citep{zerveas2021transformer}                                                          & 0.445\std{0.000}  & 0.413\std{0.000} & 0.660\std{0.000} & 0.689\std{0.000} \\
TS-SD~\citep{shi2021self}                                                        & 0.777\std{0.186}  & 0.570\std{0.033} & 0.666\std{0.044} & 0.211\std{0.000} \\
SimCLR~\citep{tang2020exploring}                                                       & 0.818\std{0.100}  & 0.422\std{0.114} & 0.496\std{0.187} & 0.471\std{0.149} \\
TS-TCC~\citep{eldele2021time}                                                       & 0.863\std{0.022}  & 0.542\std{0.034} & 0.698\std{0.036} & 0.590\std{0.095} \\
CLOCS~\citep{kiyasseh2021clocs}                                                        & 0.921\std{0.007}  & 0.475\std{0.049} & 0.401\std{0.060} & 0.514\std{0.041} \\
Mixing-up~\citep{wickstrom2022mixing}                                                    & 0.445\std{0.000}  & 0.727\std{0.023} & 0.650\std{0.031} & 0.154\std{0.020} \\
TS2Vec~\citep{yue2022ts2vec}                                                       & 0.905\std{0.007}  & 0.439\std{0.011} & 0.657\std{0.039} & 0.677\std{0.050} \\
TF-C~\citep{zhang2022self}                                                         & 0.915\std{0.053}  & 0.749\std{0.027} & 0.757\std{0.031} & 0.768\std{0.031} \\
CoST~\citep{woo2022cost}                                                         & 0.769\std{0.000}  & 0.348\std{0.000} & 0.664\std{0.000} & 0.353\std{0.000} \\
Ti-MAE~\citep{li2023ti}                                                       & 0.686\std{0.000}  & 0.666\std{0.000} & 0.684\std{0.000} & 0.709\std{0.000} \\
SimMTM~\citep{dong2024simmtm}                                                       & 0.928\std{0.000}  & 0.751\std{0.000} & 0.787\std{0.000} & \textbf{0.981\std{0.000}} \\ \midrule
\textbf{Proposed method}                                     & \textbf{0.931\std{0.003}}  & \textbf{0.867\std{0.012}} & \textbf{0.820\std{0.006}} & 0.977\std{0.017} \\ \bottomrule
\end{tabular} 
\\ \bigskip\bigskip
\small\centering\captionsetup{justification=centering, skip=5pt}
\caption{F1 score comparison between our method with a variety of learning strategies: different source datasets (\textsc{SleepEEG} and \textsc{ECG}), self-supervised learning, and random initialization (without pre-training)}
\label{tab:result2}
\begin{tabular}{@{}lC{2.1cm}C{2.1cm}C{2.1cm}C{2.1cm}@{}}
\toprule
\multicolumn{1}{c}{\textbf{Learning Strategy\hspace{1.0em}\textbackslash\hspace{1.0em}Target Dataset}} & \textbf{Epilepsy} & \textbf{FD}      & \textbf{Gesture} & \textbf{EMG}     \\ \midrule
Source: \textsc{SleepEEG}                                    & 0.931\std{0.003} & 0.867\std{0.012} & 0.820\std{0.006} & 0.977\std{0.017} \\ \midrule
Source: \textsc{ECG}                                         & 0.927\std{0.003} & 0.878\std{0.006} & 0.808\std{0.012} & 0.992\std{0.019} \\ \midrule
Self-supervised learning                                         & 0.926\std{0.002} & 0.865\std{0.008} & 0.786\std{0.006} & 0.964\std{0.011} \\ \midrule
Random initialization                                                 & 0.920\std{0.004} & 0.877\std{0.006} & 0.773\std{0.008} & 0.972\std{0.019} \\ \bottomrule
\end{tabular}
\end{table*}
\begin{figure*}[h!]
\floatconts
  {fig:ablation}
  {\caption{F1 score comparison of the proposed method with feature combinations  using $\left\{\bm{h}_t, \bm{h}_d, \bm{h}_f \right\}$} 
  }
  {%
    \includegraphics[width=0.95\linewidth]{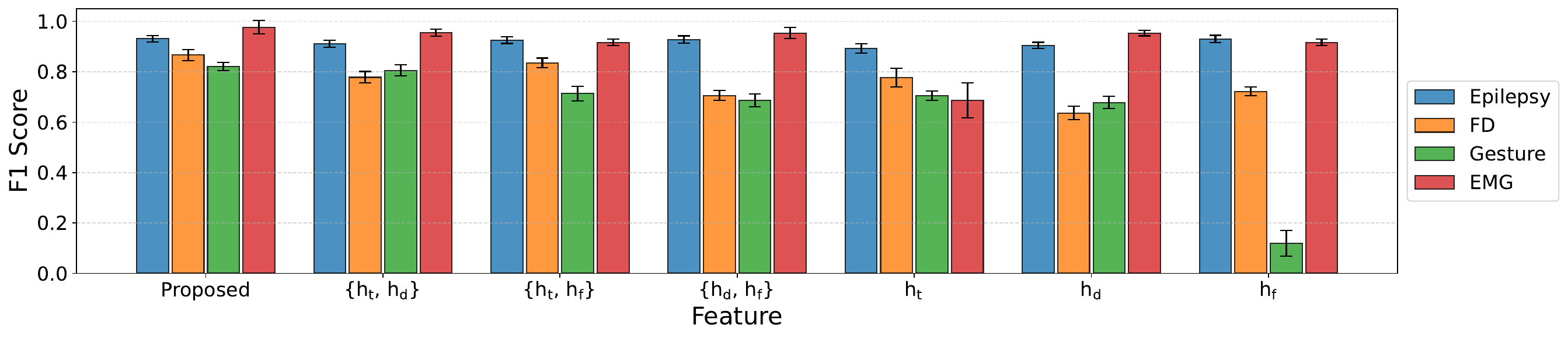}
    \vspace{-10pt}
  }
\end{figure*}

\subsection{Results with One-to-many Scenario}
In the one-to-many evaluation, the framework is pre-trained on the \textsc{SleepEEG} dataset and subsequently fine-tuned independently on multiple target datasets: \textsc{Epilepsy}, \textsc{FD}, \textsc{Gesture}, and \textsc{EMG}. This approach leverages a single, well-pretrained model from \textsc{SleepEEG}, without restarting the pre-training process for each fine-tuning task.
This evaluation tests the adaptability and generalizability of the proposed framework across datasets with diverse data types, collection protocols, and physiological or operational contexts. The fine-tuning stage involves task-specific adjustments while leveraging the feature-invariant representations learned during pre-training.

Table~\ref{tab:result1} presents the performance comparison across different methods, where the F1 score is used as the primary evaluation metric. Additional metrics, including accuracy, precision, and recall, are provided in the Appendix~\ref{appendix:results} for a more comprehensive assessment. Our method consistently outperforms the benchmark approaches, demonstrating its effectiveness in diverse experimental settings.

In Table~\ref{tab:result2}, we present an evaluation comparing different data sources, self-supervised learning approaches, and random initialization strategies. For self-supervised learning, the target dataset was utilized during the pre-training phase. In contrast, for the random initialization approach, the pre-training phase was omitted, and fine-tuning was performed directly. 
Across various scenarios, our proposed method demonstrates consistent and robust performance overall. 
Notably, the performance on the \textsc{Gesture} dataset decreases in the absence of pre-training on a larger source dataset, highlighting the importance of a substantial pre-training phase. Conversely, the \textsc{EMG} dataset achieves improved results when pre-trained with the \textsc{ECG}, suggesting that the specific characteristics of the target data play a pivotal role in performance for this case.

\figureref{fig:ablation} provides a brief comparison of performance across different feature combinations using pre-trained model using \textsc{SleepEEG}. These findings underscore the critical role of selecting features that align closely with the inherent characteristics of the target dataset, as inappropriate feature emphasis can severely impair model effectiveness. The results reveal a notable decline in performance on the \textsc{EMG} dataset when temporal features are emphasized, while the \textsc{Gesture} dataset suffers a significant drop when frequency features are prioritized. Additional experimental results and analyses across target datasets can be found in the Appendix~\ref{appendix:results}.

\section{Discussion}\label{sec:Discussion}

\paragraph{Effectiveness of proposed framework}
Our results demonstrate the substantial benefits of integrating complementary views (temporal, derivative, frequency) for robust domain adaptation. 
While our sensitivity analysis confirms that the proposed feature extraction and fusion mechanisms effectively learn transferable features, it also indicates that domain knowledge is crucial in choosing an optimal source dataset, as the best fit can vary by target task.

\paragraph{Feature Contribution Analysis}
The ablation studies provide valuable insights into the relative importance of different features. As demonstrated in \figureref{fig:ablation}, while the combination of all three feature types ($\bm{h}_t$, $\bm{h}_d$, $\bm{h}_f$) generally yields the best performance, the relative importance varies across datasets. 

\paragraph{Model Complexity}
We maintain three parallel Transformer-based encoders, one for each view. 
A standard $m$-layer Transformer has $\mathcal{O}(m\,L^2\,D)$ time complexity; as we instantiate three encoders, this adds a constant factor $\mathcal{O}(3\,\times\,m\,L^2\,D)$ plus one multi-head attention step for fusion ($\mathcal{O}(L^2\,D)$). Nevertheless, it does not alter the overall $\mathcal{O}(m\,L^2\,D)$ scaling, meaning our method remains in the same computational class as a single Transformer. Moreover, in practical scenarios $L$ is typically much larger than $m$, so the sequence-length term $L^2$ often dominates the runtime across most layers.

If $L$ is large, efficient-attention mechanisms~\citep{wang2020linformer,choromanski2021rethinking,dao2022flashattention} can reduce $\mathcal{O}(L^2)$ to $\mathcal{O}(L\log L)$ or $\mathcal{O}(L)$, thus preserving tractability for real-world time series tasks.
A promising future direction is to explore more efficient attention mechanisms and dynamic view selection to further reduce complexity.

\paragraph{Comprehensive Practical Analysis}
Despite the benefits of multi-view Transformer encoders, our design can increase computational requirement in both training and inference. To mitigate this, we employ automatic mixed precision (AMP) \citep{micikevicius2018mixed}, which reduces GPU memory requirements and accelerates large-batch contrastive pre-training. We also enable AMP at inference time, thereby improving end-to-end latency. Preliminary results are included in Appendix~\ref{appendix:computation}.

In addition to that, real-world data often feature irregularly-sampled (Appendix~\ref{appendix:irregular}) or incomplete (Appendix~\ref{appendix:partial}). We also explore the method’s applicability in multivariate time series classification beyond domain adaptation (Appendix~\ref{appendix:mtsc}). To evaluate robustness under these scenarios, we conduct additional analyses described in Appendix~\ref{appendix:comprehensive}, which provide in-depth discussions and empirical findings.

\paragraph{Note to Practitioner}
This work presents a multi-view contrastive learning framework for medical time series domain adaptation, addressing distribution shifts and temporal dependencies. 
Practitioners can leverage this methodology to enhance diagnostic and monitoring systems through robust transfer learning across diverse healthcare settings. By integrating temporal, derivative, and frequency features, the framework offers a practical solution for limited labeled data and variable collection protocols, providing actionable insights for deploying reliable machine learning models in clinical and public health contexts.

\section{Conclusion}\label{sec:Conclusion}

We proposed a novel framework for time series domain adaptation, leveraging multi-view contrastive learning to integrate complementary representations from temporal, derivative, and frequency features. By employing independent encoders and a hierarchical fusion mechanism, the framework effectively captures complex temporal dynamics and learns robust, feature-invariant representations that are transferable across domains. Extensive experiments on benchmark datasets demonstrate that the proposed method consistently outperforms state-of-the-art approaches, highlighting the benefits of multi-view feature integration for addressing domain adaptation challenges in time series analysis.

Additionally, incorporating domain-specific prior knowledge into the feature extraction process may further enhance model performance. Future work should investigate adaptive weighting mechanisms for feature fusion and explore the applicability of the framework to other types of medical time series data, including irregular or high-dimensional signals.

In the medical domain, this framework shows significant promise for improving diagnostic tools and patient monitoring systems. By learning robust representations, it addresses key challenges such as distribution shifts, data sparsity, and noise in medical time series data. The method’s ability to transfer knowledge from large pre-training datasets to smaller, specific datasets facilitates applications in scenarios with limited labeled data—a common constraint in medical research.

\acks{
We thank the teams and individuals for their efforts in dataset preparation and curation for our research. 

This research was partially supported by Basic Science Research Program through the National Research Foundation of Korea (NRF) funded by the Ministry of Education (RS-2024-00407852), and Korea Health Technology R\&D Project through the Korea Health Industry Development Institute (KHIDI), funded by the Ministry of Health and Welfare, Republic of Korea (HI19C1095).}

{\small
\bibliography{references}
}

\clearpage
\newpage
\appendix
\onecolumn

\section{Theoretical Foundations}\label{appendix:theoretical}

\paragraph{Preliminary Definitions}
Let $\mathcal{D}_S$ and $\mathcal{D}_T$ denote the {source} and {target} domain distributions over the input space $\mathcal{X}$ and label space $\mathcal{Y}$. For a hypothesis $\varphi: \mathcal{X} \to \mathcal{Y}$, its classification error on a domain $\mathcal{D}$ is:
\[
\epsilon_{\mathcal{D}}(\varphi)
\;=\;
\mathbb{E}_{(X,y)\,\sim\,\mathcal{D}}\Bigl[\,
\mathbb{I}\bigl(\varphi(X)\neq y\bigr)\Bigr].
\]
In unsupervised domain adaptation, we typically have labels only for one domain (or no labels if we do purely contrastive pre-training), and we aim to learn a predictor $\varphi$ that performs well on the target domain $\mathcal{D}_T$.
Given a sufficiently rich hypothesis class $\Phi$ (e.g., with finite VC-dimension), the classical domain-adaptation bound from \citet{ben2010theory} states:
\[
\epsilon_T(\varphi) 
\;\le\; 
\epsilon_S(\varphi) 
\;+\; 
d_{\Phi}\bigl(\mathcal{D}_S,\mathcal{D}_T\bigr)
\;+\;
\kappa,
\]
where $\kappa$ is the error of the ideal joint hypothesis on $\mathcal{D}_S \cup \mathcal{D}_T$, and
\[
d_{\Phi}(\mathcal{D}_S,\mathcal{D}_T)
\;=\;
2\,\sup_{\varphi \in \Phi}\Bigl|\,
\Pr_{X\sim \mathcal{D}_S}\bigl[\varphi(X)=1\bigr] 
\;-\;
\Pr_{X\sim \mathcal{D}_T}\bigl[\varphi(X)=1\bigr]
\Bigr|
\]
measures the distribution divergence between $\mathcal{D}_S$ and $\mathcal{D}_T$. 
In practice, we conduct contrastive pre-training on $\mathcal{D}_S$ to learn a domain-invariant encoder $E(\cdot)$ and feature mapping $F(\cdot)$ in an unsupervised manner.

\paragraph{Multi-view Contrastive Objectives}
We decompose each input $X$ into three complementary views (temporal, derivative, and frequency). Our InfoNCE loss $\mathcal{L}_{CL}$ enforces that the representation of a sample is close to its positive augmentation but dissimilar from negative samples.
For clarity, recall that the mutual information between two random variables $A$ and $B$ is defined as 
$I(A;B)=H(A)-H(A\mid B)$,
where $H(A)$ is the entropy of $A$ and $H(A\mid B)$ is the conditional entropy \citep{cover1999elements}.

\begin{proposition}[Multi-View Contrastive Learning and Mutual Information]
\label{prop:infonce_revised}
Let $\bm{z}_k$ be the feature embedding from the $k$-th view $(k\in \{t,d,f\})$, and suppose we have $K$ samples in a contrastive batch. Adapting standard results \citep{oord2018representation,Tschannen2020On}, we obtain
\[
I\bigl(\bm{z}_k ; \tilde{\bm{z}}_k\bigr)
\;\;\ge\;\;
\log K \;-\;\mathcal{L}_{CL}.
\]
Hence, minimizing $\mathcal{L}_{CL}$ increases a lower bound on the mutual information $I(\bm{z}_k; \tilde{\bm{z}}_k)$ between $\bm{z}_k$ and $\tilde{\bm{z}}_k$. In principle, as $K$ and the capacity of the encoder grow large, preserving high intra-view information fosters robust feature extraction for each view.
\end{proposition}

\paragraph{Hierarchical Multi-view Fusion}
Our {hierarchical fusion} module produces an integrated representation
\(
\bm{H}_{\mathrm{output}}
\;=\;
\bigl[\,
\bm{h}_t^*,\;\bm{h}_d^*,\;\bm{h}_f^*
\bigr],
\;=\;
\text{LayerNorm}(\bm{H} + \text{MHA}(\bm{H}))
\)
where $\bm{H} = \mathrm{stack}(\bm{h}_t,\bm{h}_d,\bm{h}_f)$.

\begin{proposition}[Inter-View Synergy via Fusion]
\label{prop:hierfusion_revised}
Under the assumption that each view $k \in \{t,d,f\}$ provides partially unique label information, the chain rule of mutual information \citep{han2021improving, hirt2024learning,oh2024deep} implies:
\[
I\bigl(\bm{H}_{\mathrm{output}};\,y\bigr)
\;\;\ge\;\;
\max\Bigl\{
I(\bm{h}_t;\,y),\;I(\bm{h}_d;\,y),\;I(\bm{h}_f;\,y)
\Bigr\}.
\]
In other words, if the temporal, derivative, and frequency views are sufficiently complementary, then fusing them can capture more label-relevant structure about $y$ than a single-view embedding alone.
\end{proposition}

\paragraph{Domain Adaptation under Multi-view Representations}
Propositions~\ref{prop:infonce_revised}--\ref{prop:hierfusion_revised} highlight how combining the three distinct feature views (temporal, derivative, frequency) can yield more label-relevant information. Consequently, domain shifts that affect only certain views can be mitigated through the complementary nature of fusion.
Suppose we have encoder $E(\cdot)$, feature extractor $F(\cdot)$, and a hierarchical feature fusion module yielding $\bm{z}_{\mathrm{combined}} = [\bm{z}_t, \bm{z}_d, \bm{z}_f]$, along with a classifier $G(\cdot)$ on top. 

\begin{theorem}[Two-Stage Multi-view Adaptation]
\label{theorem:twostage_refined}
Let $\Phi$ be a hypothesis class with finite VC-dimension, and let the source domain dataset $\mathcal{D}_S$ be sufficiently large. Assume that multi-view contrastive pre-training effectively minimizes the source embedding error $\epsilon_S(\varphi)$, producing near-optimal source representations. Furthermore, assume that each view provides complementary information about labels, thus boosting mutual information through hierarchical fusion. Then, the target-domain classification error satisfies
\[
\epsilon_T\bigl(G(\bm{z}_{\mathrm{combined}})\bigr)
\;\le\;
\epsilon_S(\varphi)
\;+\;
d_{\Phi}\bigl(\mathcal{D}_S,\mathcal{D}_T\bigr)
\;+\;
\kappa,
\]
where $\kappa$ is the minimal achievable error. Specifically, effective multi-view fusion helps reduce $d_{\Phi}(\mathcal{D}_S,\mathcal{D}_T)$ by leveraging complementary domain-invariant features across views, thereby controlling $\epsilon_T$.
\end{theorem}

If the source embedding error $\epsilon_S(\varphi)$ is minimized via multi-view contrastive pre-training, the classical domain adaptation bound \citep{ben2010theory} indicates that $\epsilon_T(\varphi)$ is mainly influenced by $d_{\Phi}(\mathcal{D}_S,\mathcal{D}_T)$ and the optimal joint error $\kappa$. By learning distinct yet complementary invariances across temporal, derivative, and frequency representations, the fusion mechanism explicitly reduces effective divergence between source and target domains. 
Realistically, $\epsilon_S(F)$ might {not} be near zero if the source data are highly variable or if contrastive pre-training is imperfect. But to the extent that multi-view contrastive learning fosters robust, domain-invariant features, it reduces $\epsilon_S(F)$ and thus can help lower $\epsilon_T(F)$.

\section{Experimental Details}\label{appendix:experimental}
The source code can be accessed at \url{https://github.com/yongkyung-oh/Multi-View_Contrastive_Learning}.
The preprocessing steps and experimental pipeline follow the methodologies described in \citet{zhang2022self}\footnote{\url{https://github.com/mims-harvard/TFC-pretraining}} and \citet{dong2024simmtm}\footnote{\url{https://github.com/thuml/SimMTM}}. For further details, please refer to their original publications.

\subsection{Datasets}

\begin{itemize}
    \item \textsc{SleepEEG}\footnote{\url{https://www.physionet.org/content/sleep-edfx/1.0.0/}}~\citep{kemp2000analysis}: This dataset contains 153 whole-night sleep electroencephalography (EEG) recordings collected from 82 healthy subjects. Each recording is sampled at 100 Hz using a 1-lead EEG signal. The EEG signals are segmented into non-overlapping windows of size 200, each forming one sample. Each sample is labeled with one of five sleep stages: Wake (W), Non-rapid Eye Movement (N1, N2, N3), and Rapid Eye Movement (REM). This segmentation results in 371,055 samples. 

    \item \textsc{ECG}\footnote{\url{https://physionet.org/content/challenge-2017/1.0.0/}}~\citep{clifford2017af}: The 2017 PhysioNet Challenge dataset focuses on classifying single-lead electrocardiogram (ECG) recordings into four classes: normal sinus rhythm, atrial fibrillation, alternative rhythm, or noisy recordings. The signals are sampled at 300 Hz and preprocessed into 5-second samples using a fixed-length window of 1,500 observations. 

    \item \textsc{Epilepsy}\footnote{\url{https://timeseriesclassification.com/description.php?Dataset=Epilepsy2}}~\citep{andrzejak2001indications}: This dataset consists of single-channel EEG recordings from 500 subjects, with each recording lasting 23.6 seconds and sampled at 178 Hz. The recordings are divided into 11,500 1-second samples, labeled according to five states: eyes open, eyes closed, healthy brain regions, tumor regions, and seizure episodes. For binary classification (seizure or non-seizure), the first four classes are merged. For fine-tuning, a subset of 60 samples (30 per class) is used, along with a validation set of 20 samples (10 per class). The remaining 11,420 samples are utilized for evaluation.

    \item \textsc{FD} (or \textsc{FD-B})\footnote{\url{https://mb.uni-paderborn.de/en/kat/main-research/datacenter/bearing-datacenter/data-sets-and-download}}~\citep{lessmeier2016condition}: This dataset, derived from an electromechanical drive system, monitor the condition of rolling bearings under varying operational conditions. The data include three classes: undamaged, inner ring damage, and outer ring damage. Original recordings (sampled at 64k Hz for 4 seconds) are processed into samples using a sliding window of 5,120 observations with sliding windows with shift lengths of 1,024 or 4,096. 

    \item \textsc{Gesture}\footnote{\url{http://www.timeseriesclassification.com/description.php?Dataset=UWaveGestureLibrary}}~\citep{liu2009uwave}: This dataset comprises accelerometer measurements capturing eight distinct hand gestures, each defined by different paths of hand movement. These gestures include swiping left, right, up, and down; waving in a counterclockwise or clockwise circle; waving in a square; and waving a right arrow. The gestures are categorized into eight classes, with 440 samples in total (55 per class), as retrieved from the UCR database. The accelerometer data, likely sampled at 100 Hz, consist of three channels representing acceleration along three coordinate axes.

    \item \textsc{EMG}\footnote{\url{https://physionet.org/content/emgdb/1.0.0/}}~\citep{goldberger2000physiobank}: This dataset consists of single-channel electromyogram (EMG) recordings from the tibialis anterior muscle of three volunteers, each diagnosed with either neuropathy or myopathy. The recordings, sampled at 4k Hz, are segmented into fixed-length samples of 1,500 observations. Each volunteer (such as, disorder type) serves as a distinct classification category.
\end{itemize}

\subsection{Benchmark methods}
\begin{itemize}
    \item \textbf{TST}~\citep{zerveas2021transformer}:
    A Transformer-based framework for multivariate time series representation learning. It utilizes self-attention mechanisms to capture long-term dependencies and demonstrates robustness across diverse datasets.
    
    \item \textbf{TS-SD}~\citep{shi2021self}:
    A self-supervised learning method for time series that employs sliding windows to generate input pairs for contrastive learning. It focuses on learning representations invariant to temporal distortions using DTW (Dynamic Time Warping).

    \item \textbf{SimCLR}~\citep{tang2020exploring}:
    Adapted from the computer vision domain \citep{chen2020simple}, SimCLR applies contrastive learning to time series by maximizing agreement between augmented views, enhancing representation quality.

    \item \textbf{TS-TCC}~\citep{eldele2021time}:
    Learns time series representations through temporal and contextual contrasting, effectively capturing both local and global temporal dynamics.

    \item \textbf{CLOCS}~\citep{kiyasseh2021clocs}:
    A contrastive learning framework for cardiac signals tailored to handle variations across space, time, and patients, with applications in clinical context.

    \item \textbf{Mixing-up}~\citep{wickstrom2022mixing}:
    Combines time series samples through interpolation to generate augmented data, improving model robustness by introducing smooth variations.

    \item \textbf{TS2Vec}~\citep{yue2022ts2vec}:
    A universal time series representation framework using hierarchical contrastive loss to capture both temporal and hierarchical structures.

    \item \textbf{TF-C}~\citep{zhang2022self}:
    Exploits temporal and frequency domain features for contrastive learning, producing robust, domain-invariant representations.

    \item \textbf{CoST}~\citep{woo2022cost}:
    Disentangles seasonal and trend representations in time series using contrastive learning, enhancing interpretability and robustness.

    \item \textbf{Ti-MAE}~\citep{li2023ti}:
    A masked autoencoder for time series data that reconstructs missing time points, learning robust and generalized representations.

    \item \textbf{SimMTM}~\citep{dong2024simmtm}:
    Leverages masked time points for representation learning through weighted aggregation of neighboring points outside the manifold, enabling generalization to time series.
\end{itemize}

\subsection{Implementation Details of the Proposed Framework}

The length of each input time series is standardized to 256 using interpolation.
The model architecture was kept consistent across the three features, $\left\{\bm{h}_t, \bm{h}_d, \bm{h}_f \right\}$. For \(k \in \{t, d, f\}\), we implemented a transformer-based encoder \(E_k\) with a hidden size of 128, 3 layers, and 4 attention heads. The feature extractor \(F_k\) was a two-layer feedforward network with \texttt{ReLU} activation, and the classifier \(G\) consisted of a single linear layer.

To mitigate overfitting and enhance generalization, we employ several complementary strategies. First, our contrastive pre-training approach encourages the learning of robust, domain-invariant representations through data augmentations and a contrastive loss objective. Second, during fine-tuning we adopt a hybrid loss function that combines the contrastive loss with cross-entropy loss, thereby balancing unsupervised representation learning with supervised classification performance. Third, we integrate standard regularization technique, such as regularization, weight decay, adaptive learning rate scheduling, and early stopping.

Due to the differing sizes of the datasets in the pre-training and fine-tuning stages, we used distinct batch sizes and epochs for each phase. For pre-training, a batch size of 128 and 200 epochs were used, while fine-tuning employed a batch size of 16 and 100 epochs. In both cases, the learning rate was set to $10^{-3}$ with a weight decay of $10^{-5}$. 
To optimize training efficiency and adaptively adjust the learning rate, we employed a learning rate scheduler. Specifically, we used a \texttt{ReduceLROnPlateau} scheduler, which monitors the validation loss and reduces the learning rate by a factor of 0.1 if the validation loss does not improve for 10 consecutive epochs. 
Additionally, we incorporated early stopping to terminate training if the validation loss did not improve for 20 consecutive epochs, further enhancing computational efficiency.

\section{Detailed Results}\label{appendix:results}

We present our experimental findings in Figure~\ref{fig:feature_combination}, which compares the F1 scores of different feature combinations pre-trained on the \textsc{SleepEEG} dataset across four target domains: \textsc{Epilepsy}, \textsc{FD}, \textsc{Gesture}, and \textsc{EMG}. Our analysis reveals that the Proposed model consistently achieves the highest F1 score across all tasks, demonstrating the effectiveness of our hierarchical fusion strategy. 
These results validate the robustness of our hierarchical fusion approach, which provides more generalizable representations across diverse tasks. However, the varying sensitivity of different tasks to individual feature combinations underscores the importance of domain expertise in optimal feature selection.

\begin{figure*}[htbp]
\floatconts
  {fig:feature_combination}
  {\caption{F1 score comparison on feature combinations of our method pre-trained on \textsc{SleepEEG} dataset}}
  {%
    \subfigure[\textsc{Epilepsy}]{\label{fig:epilepsy}%
      \includegraphics[width=0.43\linewidth]{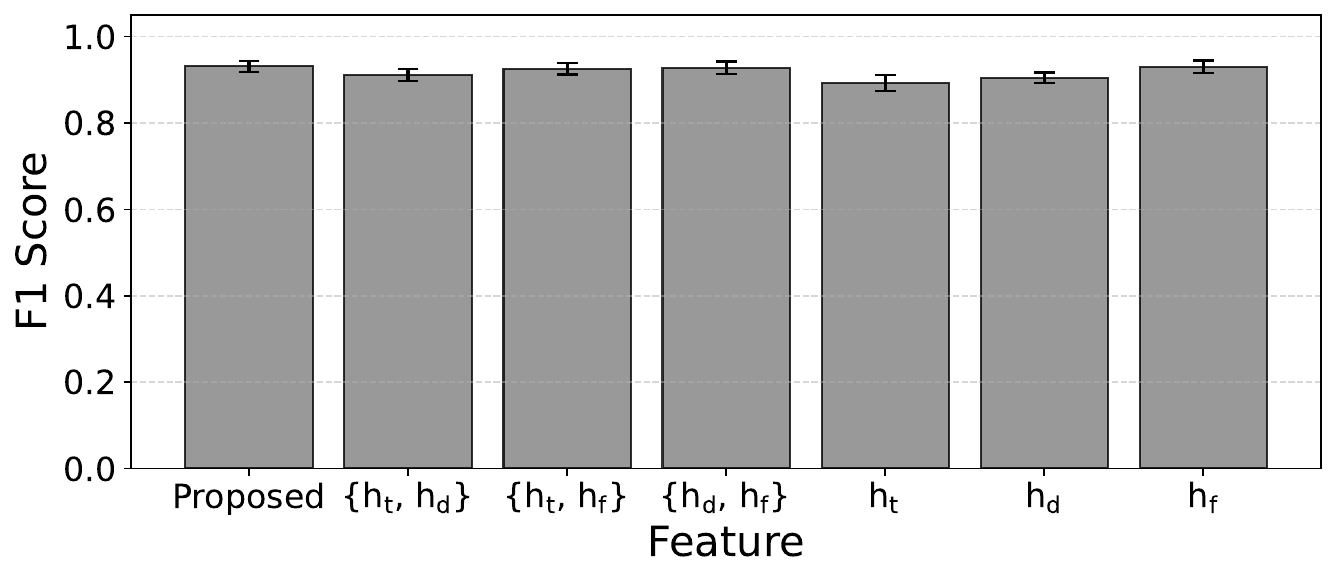}}%
    \hfil
    \subfigure[\textsc{FD}]{\label{fig:fd}%
      \includegraphics[width=0.43\linewidth]{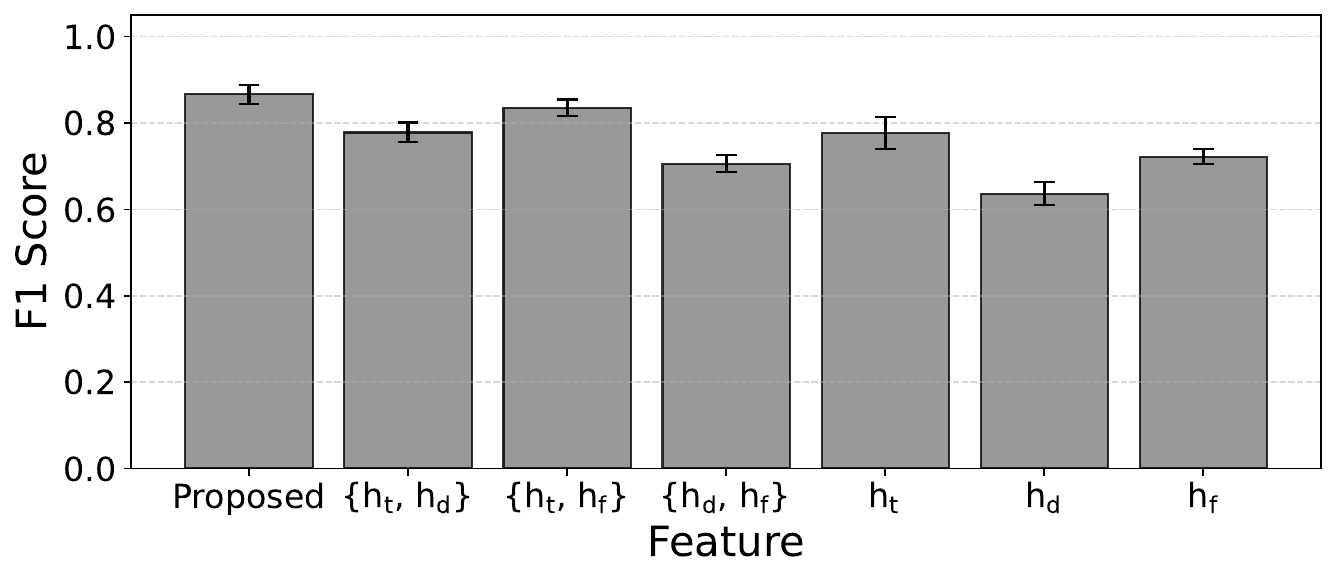}}%
    \\ \medskip    
    \subfigure[\textsc{Gesture}]{\label{fig:gesture}%
      \includegraphics[width=0.43\linewidth]{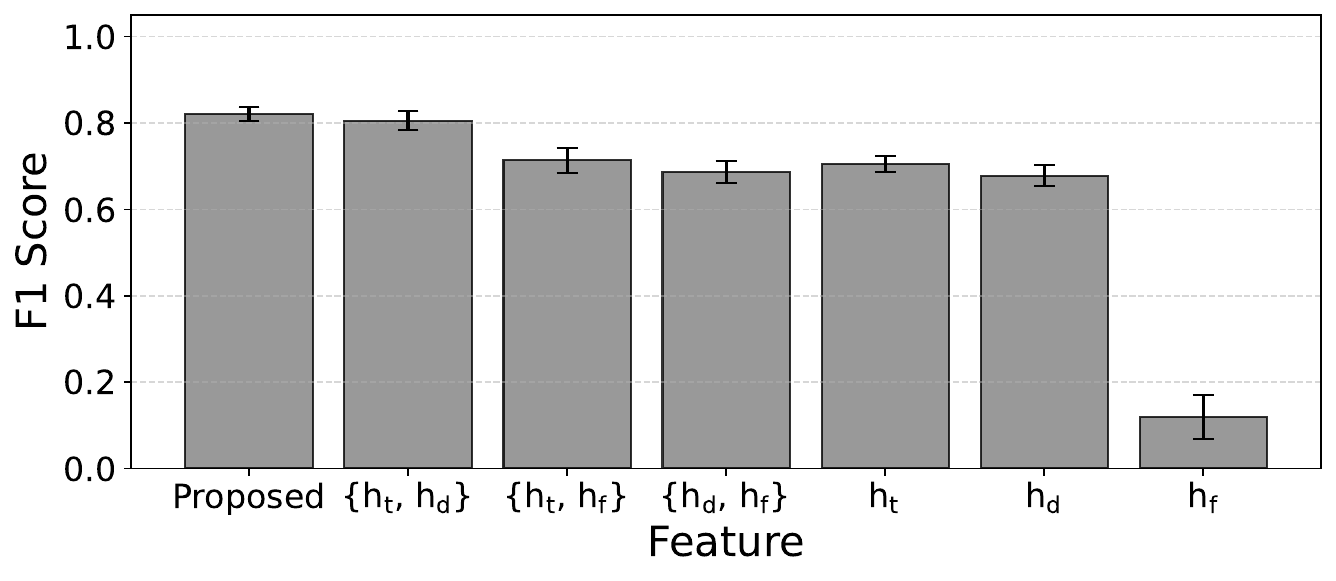}}%
    \hfil
    \subfigure[\textsc{EMG}]{\label{fig:emg}%
      \includegraphics[width=0.43\linewidth]{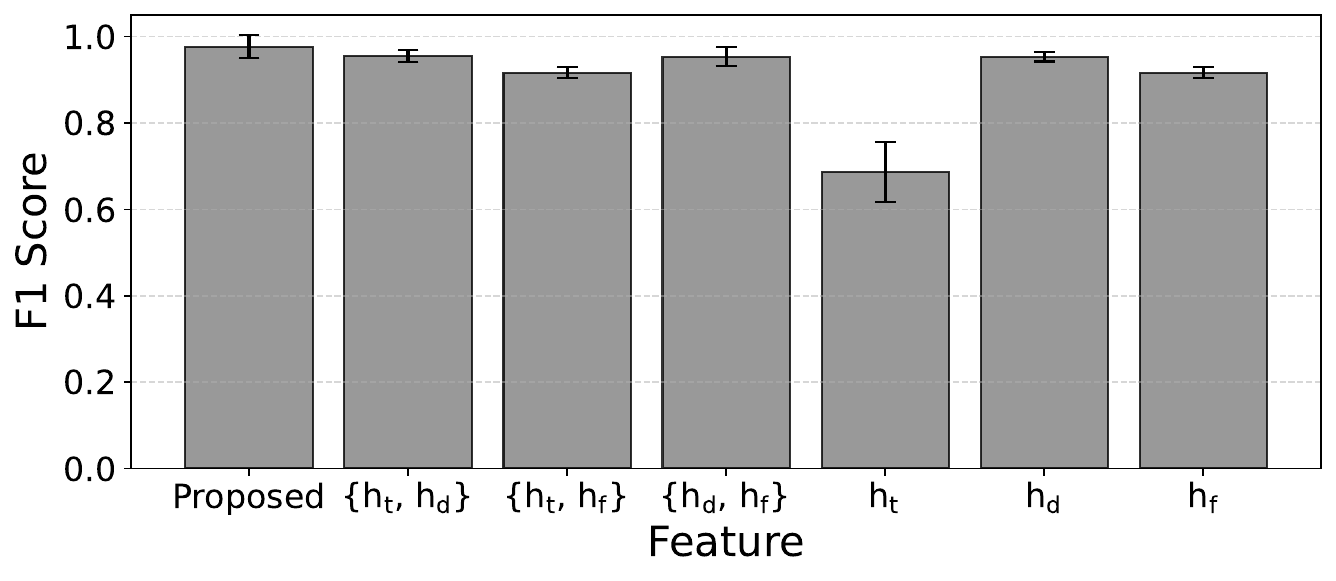}}%
    \vspace{-10pt}
  }
\end{figure*}

\begin{table*}[!t]
\small\centering\captionsetup{justification=centering, skip=5pt}
\caption{Performance comparison with benchmark method: \textsc{SleepEEG} $\rightarrow$ \textsc{FD}\\
(mean ± standard deviation over five runs; $0.000$ indicates unreported variance in the referenced studies)
}\label{tab:result2_full}
\begin{tabular}{@{}lC{2.1cm}C{2.1cm}C{2.1cm}C{2.1cm}@{}}
\toprule
\multicolumn{1}{c}{\textbf{Method\hspace{1.0em}\textbackslash\hspace{1.0em}Metric}} & \textbf{Accuracy} & \textbf{Precision} & \textbf{Recall}  & \textbf{F1 score} \\ \midrule
TST~\citep{zerveas2021transformer}                                                        & 0.464\std{0.000}  & 0.416\std{0.000}   & 0.455\std{0.000} & 0.413\std{0.000}  \\
TS-SD~\citep{shi2021self}                                                      & 0.557\std{0.021}  & 0.571\std{0.054}   & 0.605\std{0.027} & 0.570\std{0.033}  \\
SimCLR~\citep{tang2020exploring}                                                     & 0.492\std{0.044}  & 0.545\std{0.102}   & 0.476\std{0.089} & 0.422\std{0.114}  \\
TS-TCC~\citep{eldele2021time}                                                     & 0.550\std{0.022}  & 0.528\std{0.029}   & 0.640\std{0.018} & 0.542\std{0.034}  \\
CLOCS~\citep{kiyasseh2021clocs}                                                      & 0.493\std{0.031}  & 0.482\std{0.032}   & 0.587\std{0.039} & 0.475\std{0.049}  \\
Mixing-up~\citep{wickstrom2022mixing}                                                  & 0.679\std{0.025}  & 0.715\std{0.034}   & 0.761\std{0.020} & 0.727\std{0.023}  \\
TS2Vec~\citep{yue2022ts2vec}                                                     & 0.479\std{0.011}  & 0.434\std{0.009}   & 0.484\std{0.020} & 0.439\std{0.011}  \\
TF-C~\citep{zhang2022self}                                                       & 0.694\std{0.023}  & 0.756\std{0.035}   & 0.720\std{0.026} & 0.749\std{0.027}  \\
CoST~\citep{woo2022cost}                                                       & 0.471\std{0.000}  & 0.388\std{0.000}   & 0.384\std{0.000} & 0.348\std{0.000}  \\
Ti-MAE~\citep{li2023ti}                                                     & 0.609\std{0.000}  & 0.670\std{0.000}   & 0.689\std{0.000} & 0.666\std{0.000}  \\
SimMTM~\citep{dong2024simmtm}                                                     & 0.694\std{0.000}  & 0.742\std{0.000}   & 0.764\std{0.000} & 0.751\std{0.000}  \\ \midrule
\textbf{Proposed method}                                    & \textbf{0.879\std{0.010}}  & \textbf{0.853\std{0.014}}   & \textbf{0.891\std{0.009}} & \textbf{0.867\std{0.012}}  \\ \bottomrule
\end{tabular}
%
\\ \bigskip\bigskip
%
\small\centering\captionsetup{justification=centering, skip=5pt}
\caption{Sensitivity analysis on \textsc{FD} with learning strategies, including different source datasets, self-supervised learning, and random initialization 
(For each experimental scenario, classifier $G(\cdot)$ was initialized and subsequently trained by target dataset. 
`freeze scenario' indicates that $E_k(\cdot)$ and $F_k(\cdot)$ are fixed after pre-training, while only $G(\cdot)$ are optimized during fine-tuning, where $k \in \{t,d,f\}$.)
}\label{tab:result2_sensitivity}
\begin{tabular}{@{}lC{2.1cm}C{2.1cm}C{2.1cm}C{2.1cm}@{}}
\toprule
\multicolumn{1}{c}{\textbf{Learning Strategy\hspace{1.0em}\textbackslash\hspace{1.0em}Metric}} & \textbf{Accuracy} & \textbf{Precision} & \textbf{Recall}  & \textbf{F1 score} \\ \midrule
Source: \textsc{SleepEEG}                                   & 0.879\std{0.010}  & 0.853\std{0.014}   & 0.891\std{0.009} & 0.867\std{0.012}  \\
-- freeze $E_k(\cdot)$ and $F_k(\cdot)$                                        & 0.827\std{0.013}  & 0.809\std{0.011}   & 0.865\std{0.009} & 0.824\std{0.006}  \\ \midrule
Source: \textsc{ECG}                                        & 0.888\std{0.008}  & 0.862\std{0.009}   & 0.904\std{0.005} & 0.878\std{0.006}  \\
-- freeze $E_k(\cdot)$ and $F_k(\cdot)$                                        & 0.872\std{0.009}  & 0.849\std{0.014}   & 0.893\std{0.008} & 0.861\std{0.008}  \\ \midrule
Source: \textsc{FD}                                         & 0.877\std{0.008}  & 0.848\std{0.014}   & 0.889\std{0.009} & 0.865\std{0.008}  \\
-- freeze $E_k(\cdot)$ and $F_k(\cdot)$                                        & 0.869\std{0.007}  & 0.823\std{0.013}   & 0.884\std{0.005} & 0.844\std{0.011}  \\ \midrule
Random initialization                                                & 0.882\std{0.014}  & 0.866\std{0.019}   & 0.900\std{0.015} & 0.877\std{0.006}  \\ \bottomrule
\end{tabular}
%
\\ \bigskip\bigskip
%
\small\centering\captionsetup{justification=centering, skip=5pt}
\caption{Comprehensive analysis of the proposed framework's components on \textsc{FD} \\ 
(1) hierarchical feature fusion, (2) fine-tuning loss configurations with $\lambda$, and (3) combinations of features $\left\{\bm{h}_t, \bm{h}_d, \bm{h}_f \right\}$. (The proposed method utilizes all three features with $\lambda=0.1$ as the default configuration.)
}\label{tab:result2_ablation}
\begin{tabular}{@{}lC{2.1cm}C{2.1cm}C{2.1cm}C{2.1cm}@{}}
\toprule
\multicolumn{1}{c}{\textbf{Model Component\hspace{1.0em}\textbackslash\hspace{1.0em}Metric}} & \textbf{Accuracy} & \textbf{Precision} & \textbf{Recall}  & \textbf{F1 score} \\ \midrule
Proposed method                                             & 0.879\std{0.010}  & 0.853\std{0.014}   & 0.891\std{0.009} & 0.867\std{0.012}  \\
-- use   $\mathcal{L}_{CE}$ only                            & 0.879\std{0.010}  & 0.853\std{0.014}   & 0.890\std{0.010} & 0.867\std{0.012}  \\ \midrule
w/o feature   fusion                                        & 0.856\std{0.019}  & 0.838\std{0.020}   & 0.877\std{0.015} & 0.851\std{0.012}  \\
-- use   $\mathcal{L}_{CE}$ only                            & 0.847\std{0.027}  & 0.836\std{0.022}   & 0.867\std{0.021} & 0.845\std{0.017}  \\ \midrule
$\left\{\bm{h}_t,   \bm{h}_d\right\}$                       & 0.765\std{0.006}  & 0.765\std{0.018}   & 0.817\std{0.011} & 0.778\std{0.013}  \\
$\left\{\bm{h}_t,   \bm{h}_f\right\}$                       & 0.854\std{0.008}  & 0.818\std{0.018}   & 0.878\std{0.011} & 0.835\std{0.009}  \\
$\left\{\bm{h}_d,   \bm{h}_f\right\}$                       & 0.740\std{0.014}  & 0.688\std{0.011}   & 0.769\std{0.009} & 0.705\std{0.009}  \\ \midrule
$\bm{h}_t$   only                                           & 0.781\std{0.033}  & 0.755\std{0.029}   & 0.824\std{0.020} & 0.777\std{0.026}  \\
$\bm{h}_d$   only                                           & 0.671\std{0.015}  & 0.625\std{0.011}   & 0.705\std{0.011} & 0.637\std{0.017}  \\
$\bm{h}_f$   only                                           & 0.768\std{0.006}  & 0.711\std{0.008}   & 0.757\std{0.023} & 0.722\std{0.008}  \\ \bottomrule
\end{tabular}
\vspace{\fill}
\end{table*}

\begin{table*}[!t]
\small\centering\captionsetup{justification=centering, skip=5pt}
\caption{Performance comparison with benchmark methods: \textsc{SleepEEG} $\rightarrow$ \textsc{Gesture}\\
(mean ± standard deviation over five runs; $0.000$ indicates unreported variance in the referenced studies)
}\label{tab:result3_full}
\begin{tabular}{@{}lC{2.1cm}C{2.1cm}C{2.1cm}C{2.1cm}@{}}
\toprule
\multicolumn{1}{c}{\textbf{Method\hspace{1.0em}\textbackslash\hspace{1.0em}Metric}} & \textbf{Accuracy} & \textbf{Precision} & \textbf{Recall}  & \textbf{F1 score} \\ \midrule
TST~\citep{zerveas2021transformer}                                                        & 0.692\std{0.000}  & 0.666\std{0.000}   & 0.692\std{0.000} & 0.660\std{0.000}  \\
TS-SD~\citep{shi2021self}                                                      & 0.692\std{0.044}  & 0.670\std{0.047}   & 0.687\std{0.049} & 0.666\std{0.044}  \\
SimCLR~\citep{tang2020exploring}                                                     & 0.480\std{0.059}  & 0.595\std{0.162}   & 0.541\std{0.195} & 0.496\std{0.187}  \\
TS-TCC~\citep{eldele2021time}                                                     & 0.719\std{0.035}  & 0.714\std{0.035}   & 0.717\std{0.037} & 0.698\std{0.036}  \\
CLOCS~\citep{kiyasseh2021clocs}                                                      & 0.443\std{0.052}  & 0.424\std{0.079}   & 0.443\std{0.052} & 0.401\std{0.060}  \\
Mixing-up~\citep{wickstrom2022mixing}                                                  & 0.693\std{0.023}  & 0.672\std{0.023}   & 0.693\std{0.023} & 0.650\std{0.031}  \\
TS2Vec~\citep{yue2022ts2vec}                                                     & 0.692\std{0.033}  & 0.655\std{0.036}   & 0.685\std{0.035} & 0.657\std{0.039}  \\
TF-C~\citep{zhang2022self}                                                       & 0.764\std{0.020}  & 0.773\std{0.036}   & 0.743\std{0.027} & 0.757\std{0.031}  \\
CoST~\citep{woo2022cost}                                                       & 0.683\std{0.000}  & 0.653\std{0.000}   & 0.683\std{0.000} & 0.664\std{0.000}  \\
Ti-MAE~\citep{li2023ti}                                                     & 0.719\std{0.000}  & 0.704\std{0.000}   & 0.768\std{0.000} & 0.684\std{0.000}  \\
SimMTM~\citep{dong2024simmtm}                                                     & 0.800\std{0.000}  & 0.790\std{0.000}   & 0.800\std{0.000} & 0.787\std{0.000}  \\ \midrule
\textbf{Proposed method}                                    & \textbf{0.832\std{0.010}}  & \textbf{0.830\std{0.010}}   & \textbf{0.832\std{0.010}} & \textbf{0.820\std{0.006}}  \\ \bottomrule
\end{tabular}
%
\\ \bigskip\bigskip
%
\small\centering\captionsetup{justification=centering, skip=5pt}
\caption{Sensitivity analysis on \textsc{Gesture} with learning strategies, including different source datasets, self-supervised learning, and random initialization 
(For each experimental scenario, classifier $G(\cdot)$ was initialized and subsequently trained by target dataset. 
`freeze scenario' indicates that $E_k(\cdot)$ and $F_k(\cdot)$ are fixed after pre-training, while only $G(\cdot)$ are optimized during fine-tuning, where $k \in \{t,d,f\}$.)
}\label{tab:result3_sensitivity}
\begin{tabular}{@{}lC{2.1cm}C{2.1cm}C{2.1cm}C{2.1cm}@{}}
\toprule
\multicolumn{1}{c}{\textbf{Learning Strategy\hspace{1.0em}\textbackslash\hspace{1.0em}Metric}} & \textbf{Accuracy} & \textbf{Precision} & \textbf{Recall}  & \textbf{F1 score} \\ \midrule
Source: \textsc{SleepEEG}                                   & 0.832\std{0.010}  & 0.830\std{0.010}   & 0.832\std{0.010} & 0.820\std{0.006}  \\
-- freeze $E_k(\cdot)$ and $F_k(\cdot)$                                        & 0.833\std{0.007}  & 0.832\std{0.008}   & 0.833\std{0.007} & 0.820\std{0.006}  \\ \midrule
Source: \textsc{ECG}                                        & 0.817\std{0.011}  & 0.815\std{0.019}   & 0.817\std{0.011} & 0.808\std{0.012}  \\
-- freeze $E_k(\cdot)$ and $F_k(\cdot)$                                        & 0.735\std{0.008}  & 0.740\std{0.006}   & 0.735\std{0.008} & 0.719\std{0.008}  \\ \midrule
Source: \textsc{Gesture}                                    & 0.792\std{0.007}  & 0.788\std{0.008}   & 0.792\std{0.007} & 0.786\std{0.006}  \\
-- freeze $E_k(\cdot)$ and $F_k(\cdot)$                                        & 0.792\std{0.007}  & 0.788\std{0.008}   & 0.792\std{0.007} & 0.786\std{0.006}  \\ \midrule
Random initialization                                                & 0.790\std{0.010}  & 0.779\std{0.015}   & 0.790\std{0.010} & 0.773\std{0.008}  \\ \bottomrule
\end{tabular}
%
\\ \bigskip\bigskip
%
\small\centering\captionsetup{justification=centering, skip=5pt}
\caption{Comprehensive analysis of the proposed framework's components on \textsc{Gesture} \\ 
(1) hierarchical feature fusion, (2) fine-tuning loss configurations with $\lambda$, and (3) combinations of features $\left\{\bm{h}_t, \bm{h}_d, \bm{h}_f \right\}$. (The proposed method utilizes all three features with $\lambda=0.1$ as the default configuration.)
}\label{tab:result3_ablation}
\begin{tabular}{@{}lC{2.1cm}C{2.1cm}C{2.1cm}C{2.1cm}@{}}
\toprule
\multicolumn{1}{c}{\textbf{Model Component\hspace{1.0em}\textbackslash\hspace{1.0em}Metric}} & \textbf{Accuracy} & \textbf{Precision} & \textbf{Recall}  & \textbf{F1 score} \\ \midrule
Proposed method                                             & 0.832\std{0.010}  & 0.830\std{0.010}   & 0.832\std{0.010} & 0.820\std{0.006}  \\
-- use   $\mathcal{L}_{CE}$ only                            & 0.833\std{0.007}  & 0.832\std{0.008}   & 0.833\std{0.007} & 0.820\std{0.006}  \\ \midrule
w/o feature   fusion                                        & 0.808\std{0.007}  & 0.807\std{0.006}   & 0.808\std{0.007} & 0.798\std{0.011}  \\
-- use   $\mathcal{L}_{CE}$ only                            & 0.798\std{0.012}  & 0.791\std{0.016}   & 0.798\std{0.012} & 0.785\std{0.012}  \\ \midrule
$\left\{\bm{h}_t,   \bm{h}_d\right\}$                       & 0.815\std{0.012}  & 0.813\std{0.008}   & 0.815\std{0.012} & 0.805\std{0.012}  \\
$\left\{\bm{h}_t,   \bm{h}_f\right\}$                       & 0.743\std{0.016}  & 0.729\std{0.034}   & 0.743\std{0.016} & 0.714\std{0.019}  \\
$\left\{\bm{h}_d,   \bm{h}_f\right\}$                       & 0.698\std{0.019}  & 0.711\std{0.024}   & 0.698\std{0.019} & 0.687\std{0.016}  \\ \midrule
$\bm{h}_t$   only                                           & 0.732\std{0.012}  & 0.726\std{0.032}   & 0.732\std{0.012} & 0.705\std{0.008}  \\
$\bm{h}_d$   only                                           & 0.690\std{0.016}  & 0.697\std{0.025}   & 0.690\std{0.016} & 0.678\std{0.014}  \\
$\bm{h}_f$   only                                           & 0.202\std{0.033}  & 0.122\std{0.066}   & 0.202\std{0.033} & 0.119\std{0.042}  \\ \bottomrule
\end{tabular}
\vspace{\fill}
\end{table*}

\begin{table*}[!t]
\small\centering\captionsetup{justification=centering, skip=5pt}
\caption{Performance comparison with benchmark methods: \textsc{SleepEEG} $\rightarrow$ \textsc{EMG}\\
(mean ± standard deviation over five runs; $0.000$ indicates unreported variance in the referenced studies)
}\label{tab:result4_full}
\begin{tabular}{@{}lC{2.1cm}C{2.1cm}C{2.1cm}C{2.1cm}@{}}
\toprule
\multicolumn{1}{c}{\textbf{Method\hspace{1.0em}\textbackslash\hspace{1.0em}Metric}} & \textbf{Accuracy} & \textbf{Precision} & \textbf{Recall}  & \textbf{F1 score} \\ \midrule
TST~\citep{zerveas2021transformer}                                                        & 0.783\std{0.000}  & 0.771\std{0.000}   & 0.803\std{0.000} & 0.689\std{0.000}  \\
TS-SD~\citep{shi2021self}                                                      & 0.461\std{0.000}  & 0.155\std{0.000}   & 0.333\std{0.000} & 0.211\std{0.000}  \\
SimCLR~\citep{tang2020exploring}                                                     & 0.615\std{0.058}  & 0.516\std{0.172}   & 0.499\std{0.121} & 0.471\std{0.149}  \\
TS-TCC~\citep{eldele2021time}                                                     & 0.789\std{0.019}  & 0.585\std{0.097}   & 0.631\std{0.099} & 0.590\std{0.095}  \\
CLOCS~\citep{kiyasseh2021clocs}                                                      & 0.699\std{0.032}  & 0.511\std{0.075}   & 0.515\std{0.029} & 0.514\std{0.041}  \\
Mixing-up~\citep{wickstrom2022mixing}                                                  & 0.302\std{0.053}  & 0.110\std{0.013}   & 0.258\std{0.046} & 0.154\std{0.020}  \\
TS2Vec~\citep{yue2022ts2vec}                                                     & 0.785\std{0.032}  & 0.804\std{0.075}   & 0.679\std{0.040} & 0.677\std{0.050}  \\
TF-C~\citep{zhang2022self}                                                       & 0.817\std{0.029}  & 0.727\std{0.035}   & 0.816\std{0.029} & 0.768\std{0.031}  \\
CoST~\citep{woo2022cost}                                                       & 0.517\std{0.000}  & 0.491\std{0.000}   & 0.421\std{0.000} & 0.353\std{0.000}  \\
Ti-MAE~\citep{li2023ti}                                                     & 0.700\std{0.000}  & 0.703\std{0.000}   & 0.634\std{0.000} & 0.709\std{0.000}  \\
SimMTM~\citep{dong2024simmtm}                                                     & 0.976\std{0.000}  & \textbf{0.983\std{0.000}}   & 0.980\std{0.000} & \textbf{0.981\std{0.000}}  \\ \midrule
\textbf{Proposed method}                                    & \textbf{0.980\std{0.012}}  & 0.976\std{0.022}   & \textbf{0.984\std{0.010}} & 0.977\std{0.017}  \\ \bottomrule
\end{tabular}
%
\\ \bigskip\bigskip
%
\small\centering\captionsetup{justification=centering, skip=5pt}
\caption{Sensitivity analysis on \textsc{EMG} with learning strategies, including different source datasets, self-supervised learning, and random initialization 
(For each experimental scenario, classifier $G(\cdot)$ was initialized and subsequently trained by target dataset. 
`freeze scenario' indicates that $E_k(\cdot)$ and $F_k(\cdot)$ are fixed after pre-training, while only $G(\cdot)$ are optimized during fine-tuning, where $k \in \{t,d,f\}$.)
}\label{tab:result4_sensitivity}
\begin{tabular}{@{}lC{2.1cm}C{2.1cm}C{2.1cm}C{2.1cm}@{}}
\toprule
\multicolumn{1}{c}{\textbf{Learning Strategy\hspace{1.0em}\textbackslash\hspace{1.0em}Metric}} & \textbf{Accuracy} & \textbf{Precision} & \textbf{Recall}  & \textbf{F1 score} \\ \midrule
Source: \textsc{SleepEEG}                                   & 0.980\std{0.012}  & 0.976\std{0.022}   & 0.984\std{0.010} & 0.977\std{0.017}  \\
-- freeze $E_k(\cdot)$ and $F_k(\cdot)$                                        & 0.971\std{0.012}  & 0.973\std{0.018}   & 0.976\std{0.010} & 0.973\std{0.012}  \\ \midrule
Source: \textsc{ECG}                                        & 0.995\std{0.012}  & 0.997\std{0.008}   & 0.996\std{0.010} & 0.992\std{0.019}  \\
-- freeze $E_k(\cdot)$ and $F_k(\cdot)$                                        & 0.995\std{0.012}  & 0.997\std{0.008}   & 0.992\std{0.019} & 0.991\std{0.021}  \\ \midrule
Source: \textsc{EMG}                                        & 0.976\std{0.001}  & 0.974\std{0.018}   & 0.980\std{0.001} & 0.964\std{0.011}  \\
-- freeze $E_k(\cdot)$ and $F_k(\cdot)$                                        & 0.976\std{0.001}  & 0.967\std{0.021}   & 0.971\std{0.022} & 0.958\std{0.004}  \\ \midrule
Random initialization                                                & 0.980\std{0.012}  & 0.978\std{0.021}   & 0.984\std{0.010} & 0.972\std{0.019}  \\ \bottomrule
\end{tabular}
%
\\ \bigskip\bigskip
%
\small\centering\captionsetup{justification=centering, skip=5pt}
\caption{Comprehensive analysis of the proposed framework's components on \textsc{EMG} \\ 
(1) hierarchical feature fusion, (2) fine-tuning loss configurations with $\lambda$, and (3) combinations of features $\left\{\bm{h}_t, \bm{h}_d, \bm{h}_f \right\}$. (The proposed method utilizes all three features with $\lambda=0.1$ as the default configuration.)
}\label{tab:result4_ablation}
\begin{tabular}{@{}lC{2.1cm}C{2.1cm}C{2.1cm}C{2.1cm}@{}}
\toprule
\multicolumn{1}{c}{\textbf{Model Component\hspace{1.0em}\textbackslash\hspace{1.0em}Metric}} & \textbf{Accuracy} & \textbf{Precision} & \textbf{Recall}  & \textbf{F1 score} \\ \midrule
Proposed method                                             & 0.980\std{0.012}  & 0.976\std{0.022}   & 0.984\std{0.010} & 0.977\std{0.017}  \\
-- use   $\mathcal{L}_{CE}$ only                            & 0.976\std{0.001}  & 0.973\std{0.018}   & 0.980\std{0.001} & 0.973\std{0.012}  \\ \midrule
w/o feature fusion                                          & 0.971\std{0.012}  & 0.973\std{0.018}   & 0.976\std{0.010} & 0.973\std{0.012}  \\
-- use   $\mathcal{L}_{CE}$ only                            & 0.966\std{0.014}  & 0.969\std{0.025}   & 0.973\std{0.012} & 0.969\std{0.019}  \\ \midrule
$\left\{\bm{h}_t,   \bm{h}_d\right\}$                       & 0.976\std{0.001}  & 0.981\std{0.001}   & 0.943\std{0.022} & 0.955\std{0.004}  \\
$\left\{\bm{h}_t,   \bm{h}_f\right\}$                       & 0.941\std{0.014}  & 0.929\std{0.018}   & 0.912\std{0.010} & 0.916\std{0.003}  \\
$\left\{\bm{h}_d,   \bm{h}_f\right\}$                       & 0.971\std{0.012}  & 0.956\std{0.019}   & 0.958\std{0.032} & 0.954\std{0.011}  \\ \midrule
$\bm{h}_t$   only                                           & 0.873\std{0.012}  & 0.754\std{0.163}   & 0.698\std{0.034} & 0.686\std{0.060}  \\
$\bm{h}_d$   only                                           & 0.976\std{0.001}  & 0.981\std{0.001}   & 0.933\std{0.001} & 0.953\std{0.001}  \\
$\bm{h}_f$   only                                           & 0.941\std{0.014}  & 0.929\std{0.018}   & 0.907\std{0.011} & 0.916\std{0.003}  \\ \bottomrule
\end{tabular}
\vspace{\fill}
\end{table*}

\paragraph{\textsc{FD}} Table~\ref{tab:result2_full} shows the proposed method outperforming benchmarks on the \textsc{FD} dataset, showcasing its adaptability to industrial time series. Table~\ref{tab:result2_sensitivity} highlights the robustness of learning strategies, and Table~\ref{tab:result2_ablation} confirms the significant impact of multi-view fusion over single features.

\paragraph{\textsc{Gesture}} Table~\ref{tab:result3_full} demonstrates state-of-the-art performance by the proposed method on the \textsc{Gesture} dataset. Table~\ref{tab:result3_sensitivity} validates \textsc{SleepEEG} as the best pre-training source, while Table~\ref{tab:result3_ablation} emphasizes the importance of hierarchical feature fusion.

\paragraph{\textsc{EMG}} Table~\ref{tab:result4_full} highlights the superior performance of the proposed method on the \textsc{EMG} dataset. Table~\ref{tab:result4_sensitivity} confirms the value of fine-tuning on pre-trained models, and Table~\ref{tab:result4_ablation} showcase the importance of each feature. Notably, the self-supervised learning approach achieves superior performance compared to other.

\medskip

It is important to note that the optimal source dataset varies depending on the target domain, highlighting the significance of incorporating domain knowledge and sophisticated feature selection approach.

\section{Comprehensive Practical Analysis}\label{appendix:comprehensive}

\subsection{Practical Computation Strategy}\label{appendix:computation}
Table~\ref{tab:amp_sensitivity} shows that half-precision arithmetic by AMP~\citep{micikevicius2018mixed} attains competitive classification metrics across different pre-training sources on \textsc{Epilepsy}. 
Moreover, we observed an average batch inference time of $0.0519 \pm 0.0002\,\mathrm{s}$ with AMP versus $0.0823 \pm 0.0003\,\mathrm{s}$ in full precision mode, indicating a tangible speedup. 
Note that these computation times may vary depending on hardware specifications, batch sizes, and other implementation details. 
Even so, AMP can reduce memory usage and computational cost during training while also accelerating inference, helping to offset the additional overhead associated with multi-view feature extraction and still maintaining robust domain adaptation performance.

\begin{table*}[!htb]
\small\centering\captionsetup{justification=centering, skip=5pt}
\caption{Comparative analysis of performance on \textsc{Epilepsy} using three different source datasets (\textsc{SleepEEG}, and \textsc{ECG}) under automatic mixed precision (AMP) vs.\ full-precision settings. 
\label{tab:amp_sensitivity}
}
\begin{tabular}{@{}clC{2.1cm}C{2.1cm}C{2.1cm}C{2.1cm}@{}}
\toprule
\textbf{Setting}            & \multicolumn{1}{c}{\textbf{Domain Adaptation}} & \textbf{Accuracy} & \textbf{Precision} & \textbf{Recall}  & \textbf{F1 score} \\ \midrule
\multirow{2}{*}{with   AMP} &  \textsc{SleepEEG} $\rightarrow$ \textsc{Epilepsy}                                & 0.958\std{0.002}  & 0.936\std{0.004}   & 0.935\std{0.004} & 0.931\std{0.003}  \\
                            &  \textsc{ECG} $\rightarrow$ \textsc{Epilepsy}                                      & 0.956\std{0.002}  & 0.934\std{0.004}   & 0.940\std{0.008} & 0.927\std{0.003}  \\ \midrule
\multirow{2}{*}{full-precision}  &  \textsc{SleepEEG} $\rightarrow$ \textsc{Epilepsy}                                 & 0.957\std{0.003}  & 0.939\std{0.014}   & 0.942\std{0.004} & 0.932\std{0.004}  \\
                            &  \textsc{ECG} $\rightarrow$ \textsc{Epilepsy}                                      & 0.955\std{0.004}  & 0.934\std{0.010}   & 0.944\std{0.005} & 0.928\std{0.005}  \\ \bottomrule
\end{tabular}
\end{table*}

\subsection{Handling Irregularly-Sampled Time Series Data}\label{appendix:irregular}
Real-world medical or sensor data are frequently characterized by irregular sampling intervals and missing values. A straightforward way to address this in our framework is to use the same interpolation mechanism adopted for derivative computation. 
%
Figure~\ref{fig:irregular_epilepsy}~(a) shows the original time series, (b) illustrates an artificially introduced missing scenario where we randomly remove about 50\% of observations from each sample, independently, leading to irregular sampling.
Figure~\ref{fig:irregular_epilepsy}~(c) and (d) depict the spline-based interpolation used to recover a uniform grid. Although interpolation closely aligns with the original data, unavoidable deviations from missing observations introduce a dataset shift.

Furthermore, Figure~\ref{fig:feature_missing} compares the extracted features (temporal, derivative, frequency) obtained from the original versus interpolated signals from the selected samples. The black curves indicate features from the original data, while the red curves show those derived via interpolation-based resampling. In practice, this interpolation preserves the core patterns in all three feature views.

Table~\ref{tab:missing_sensitivity} reports classification for two scenarios: using the original data, and using an interpolation-based method to handle irregular or missing observations. We pre-trained the model on regularly sampled time series but performed fine-tuning with irregularly-sampled data. Despite a slight performance drop in the interpolated setting, the overall metrics remain feasible, indicating that our framework is robust even with irregularly sampling or missing data. 
In future work, we intend to explore more sophisticated methods for managing irregularly-sampled time series (e.g., continuous-time models or neural differential equations) and further enrich our framework’s ability to handle incomplete real-world data.

\begin{figure*}[htb]
\floatconts
{fig:irregular_epilepsy}
{\caption{Example of handling irregularly-sampled observations in the \textsc{Epilepsy} dataset}}
{%
    \subfigure[Original observation]{\label{fig:orig}%
      \includegraphics[width=0.42\linewidth]{figs/missing/Epilepsy.pdf}}%
    \hfil
    \subfigure[50\% missing scenario]{\label{fig:missing}%
      \includegraphics[width=0.42\linewidth]{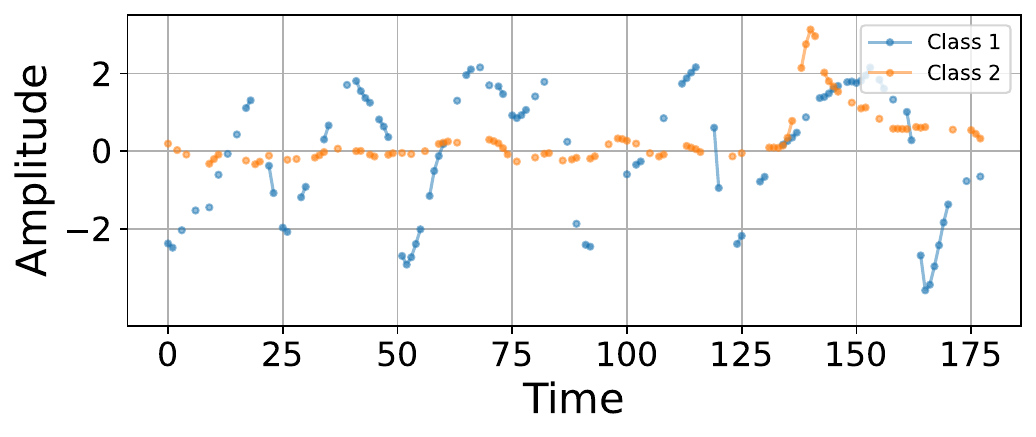}}%
    \\ \medskip
    \subfigure[Interpolation (Class 1)]{\label{fig:interp1}%
      \includegraphics[width=0.42\linewidth]{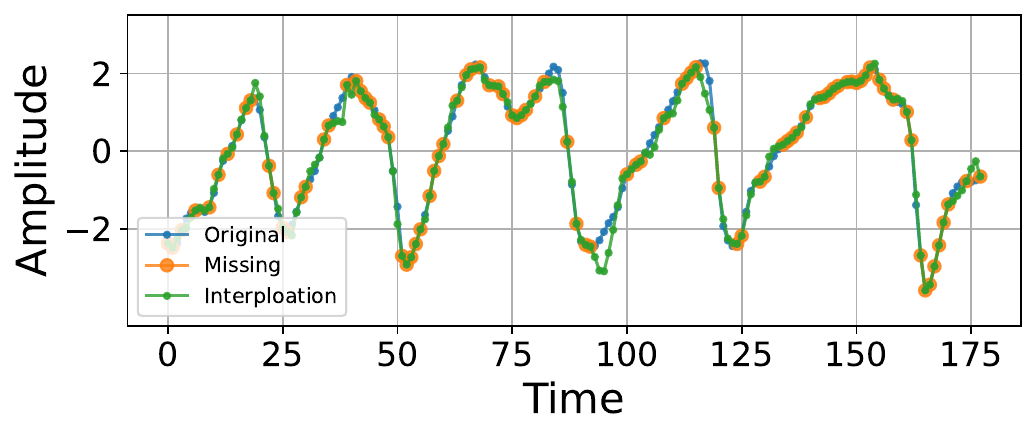}}%
    \hfil
    \subfigure[Interpolation (Class 2)]{\label{fig:interp2}%
      \includegraphics[width=0.42\linewidth]{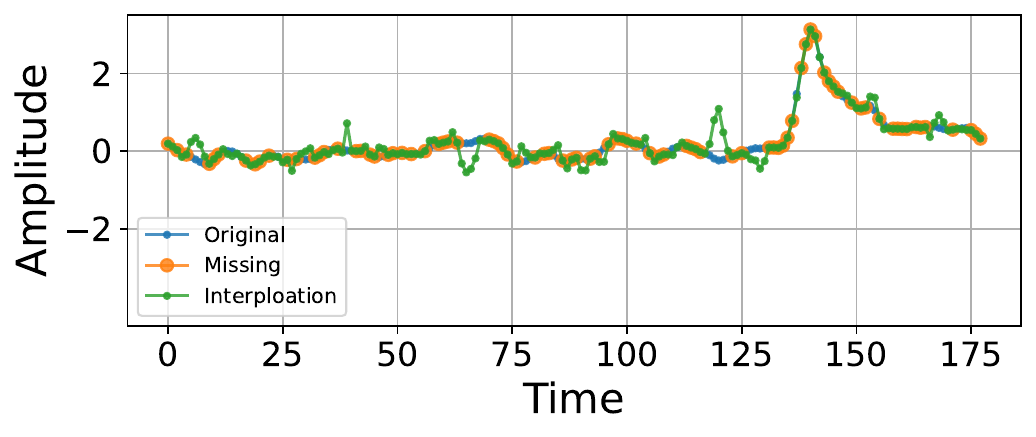}}%
    \vspace{-10pt}
}
\end{figure*}
\begin{figure*}[htb]
\floatconts
  {fig:feature_missing}
  {\caption{Comparison of features computed on the original versus interpolated \textsc{Epilepsy} dataset}}
  {%
    \subfigure[Temporal (Class 1)]{\label{fig:temporal1}%
      \includegraphics[width=0.32\linewidth]{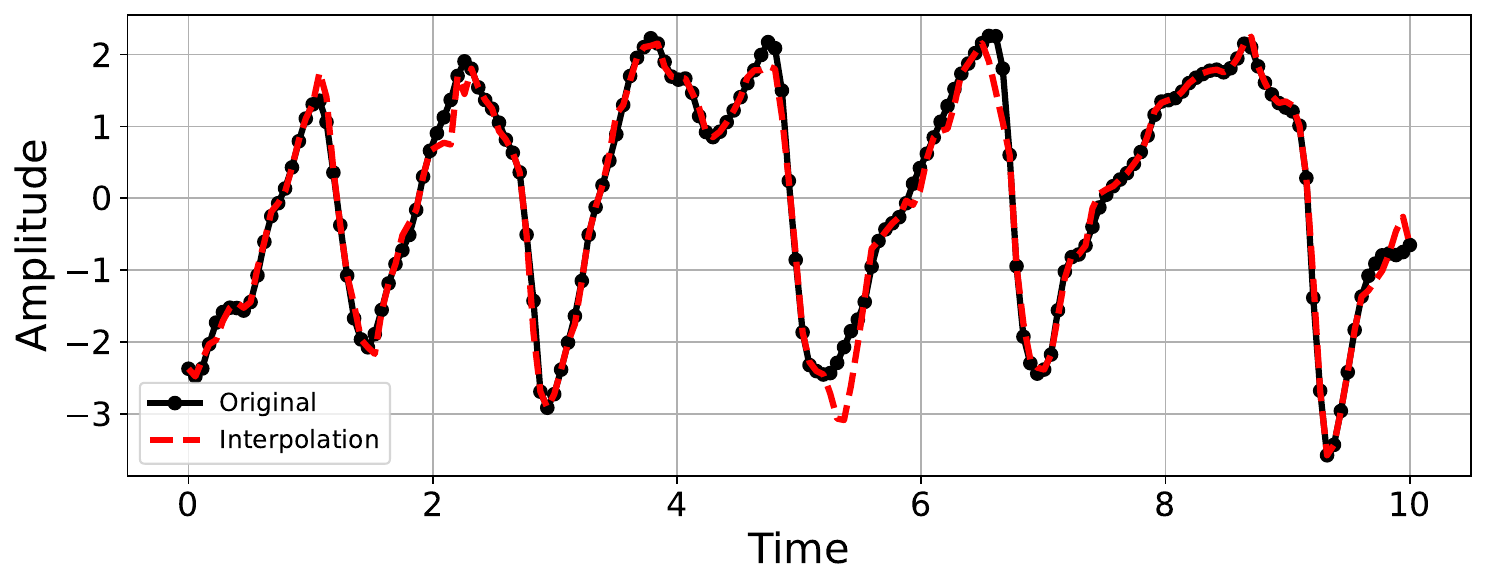}}%
    \hfil
    \subfigure[Derivative (Class 1)]{\label{fig:derivative1}%
      \includegraphics[width=0.32\linewidth]{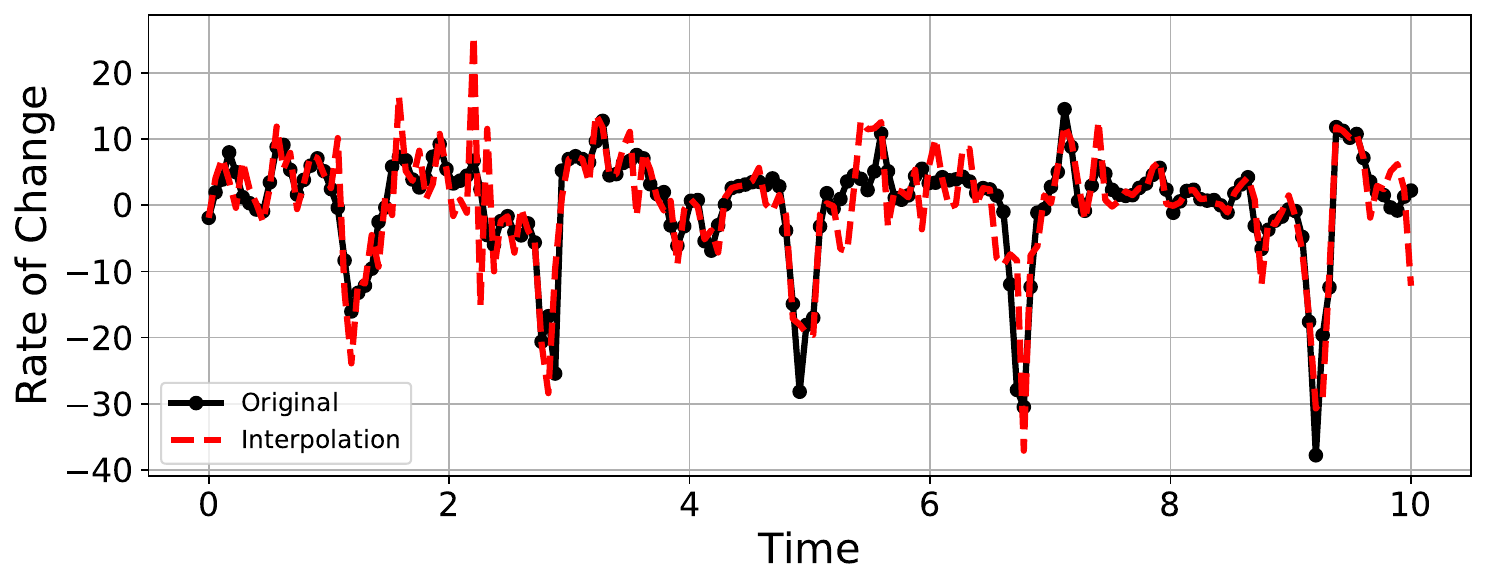}}%
    \hfil
    \subfigure[Frequency (Class 1)]{\label{fig:frequency1}%
      \includegraphics[width=0.32\linewidth]{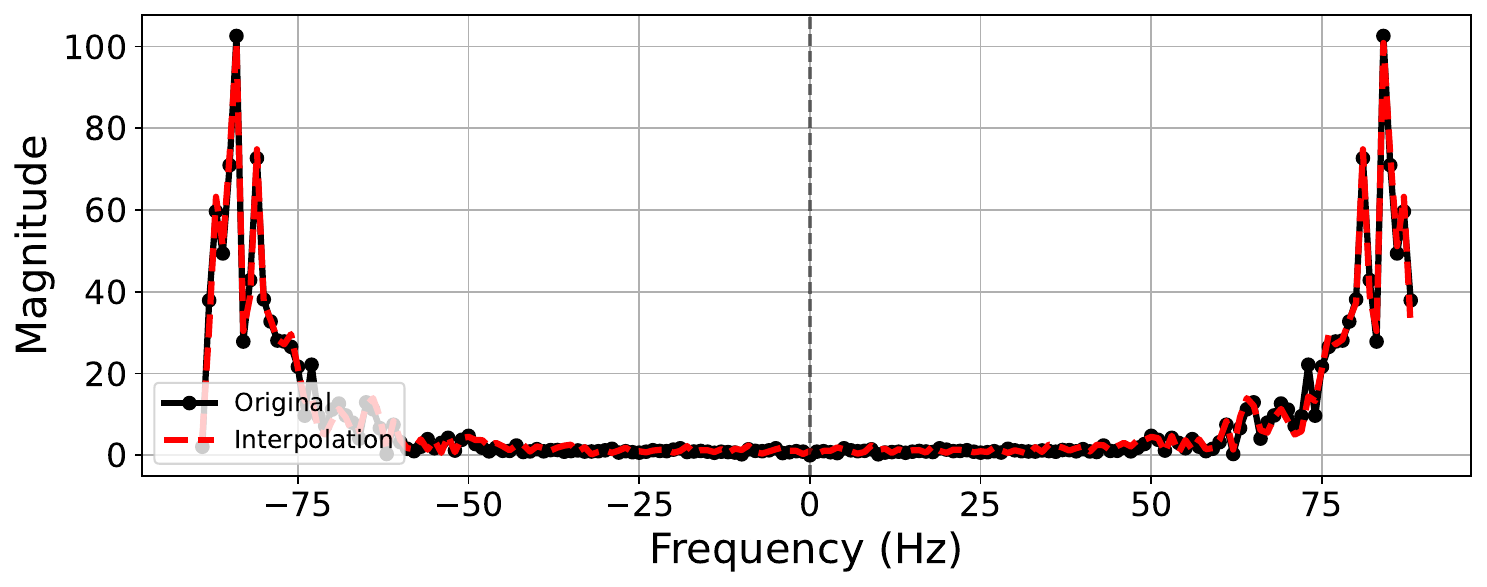}}%
    \\ \medskip
    \subfigure[Temporal (Class 2)]{\label{fig:temporal2}%
      \includegraphics[width=0.32\linewidth]{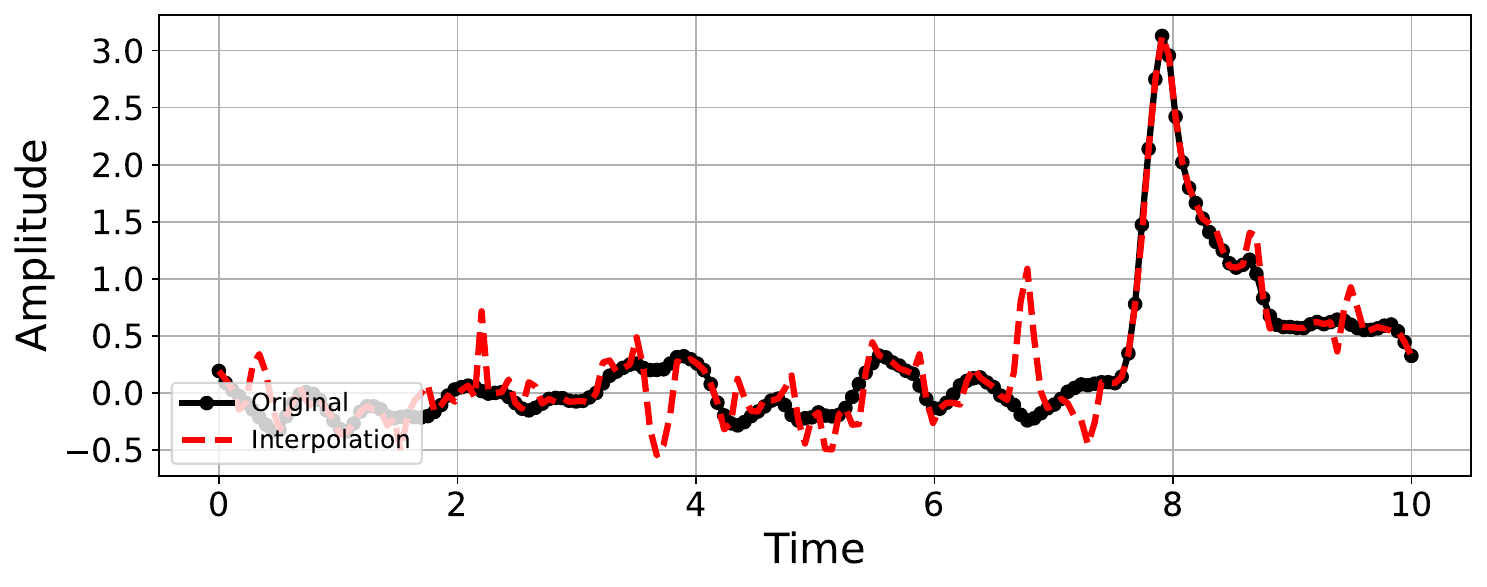}}%
    \hfil
    \subfigure[Derivative (Class 2)]{\label{fig:derivative2}%
      \includegraphics[width=0.32\linewidth]{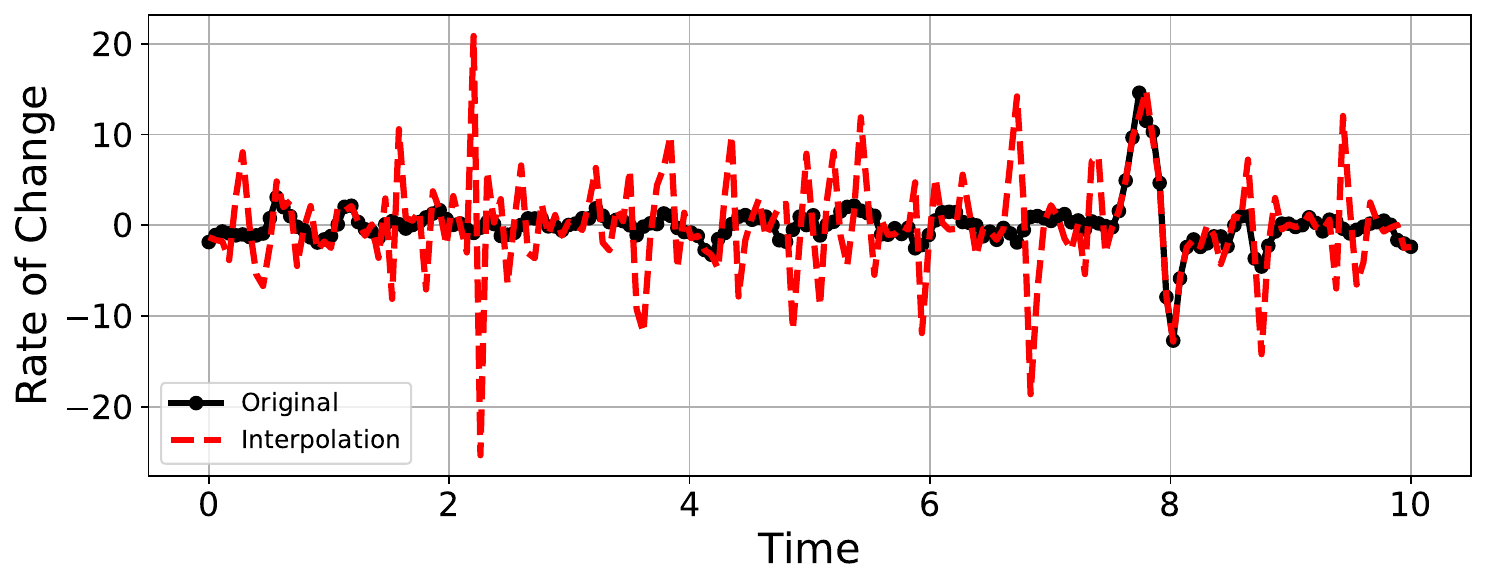}}%
    \hfil
    \subfigure[Frequency (Class 2)]{\label{fig:frequency2}%
      \includegraphics[width=0.32\linewidth]{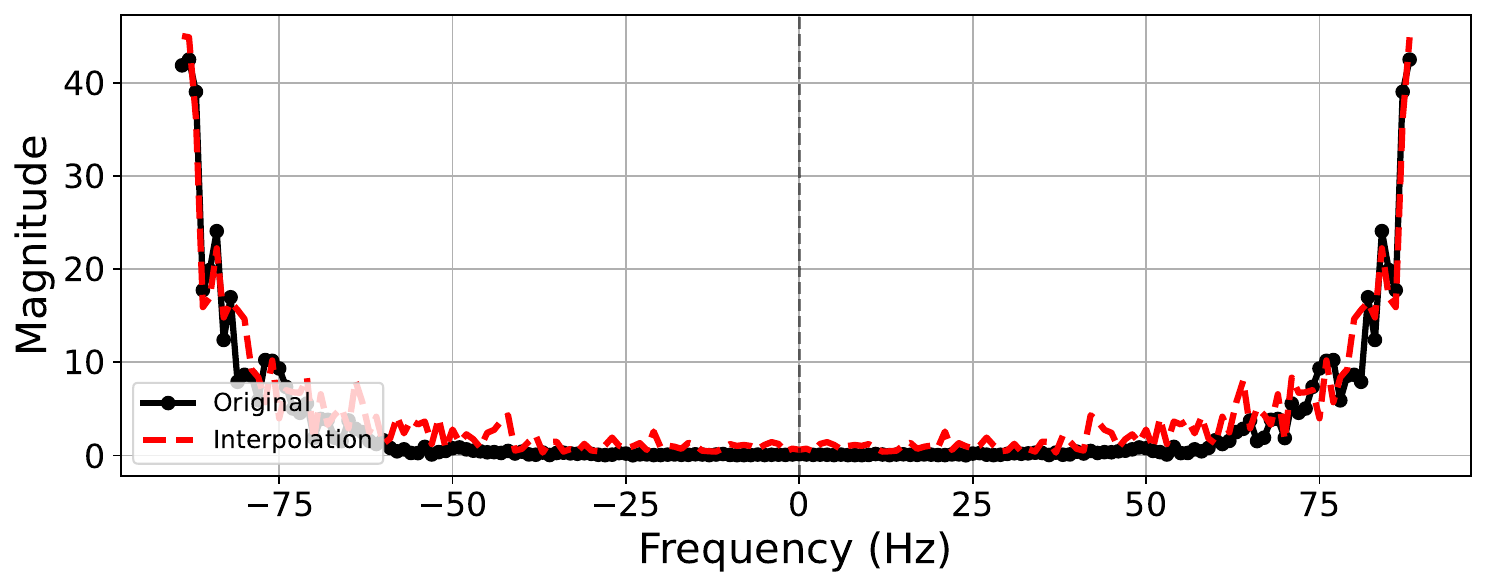}}%
    \vspace{-10pt}  
  }
\end{figure*}

\begin{table*}[!htb]
\small\centering\captionsetup{justification=centering, skip=5pt}
\caption{Comparative analysis of performance on \textsc{Epilepsy} with irregularly-sampled scenario
\label{tab:missing_sensitivity}
}
\begin{tabular}{@{}clC{2.1cm}C{2.1cm}C{2.1cm}C{2.1cm}@{}}
\toprule
\textbf{Setting}            & \multicolumn{1}{c}{\textbf{Domain Adaptation}} & \textbf{Accuracy} & \textbf{Precision} & \textbf{Recall}  & \textbf{F1 score} \\ \midrule
\multirow{2}{*}{Original}      &  \textsc{SleepEEG} $\rightarrow$ \textsc{Epilepsy}                                 & 0.958\std{0.002}  & 0.936\std{0.004}   & 0.935\std{0.004} & 0.931\std{0.003}  \\
                               &  \textsc{ECG} $\rightarrow$ \textsc{Epilepsy}                                      & 0.956\std{0.002}  & 0.934\std{0.004}   & 0.940\std{0.008} & 0.927\std{0.003}  \\ \midrule
\multirow{2}{*}{Interpolation} &  \textsc{SleepEEG} $\rightarrow$ \textsc{Epilepsy}                                 & 0.951\std{0.003}  & 0.937\std{0.013}   & 0.929\std{0.005} & 0.922\std{0.003}  \\
                               &  \textsc{ECG} $\rightarrow$ \textsc{Epilepsy}                                      & 0.951\std{0.003}  & 0.927\std{0.007}   & 0.931\std{0.004} & 0.923\std{0.004}  \\ \bottomrule
\end{tabular}
\end{table*}

\subsection{Impact of Source Dataset in Domain Adaptation}\label{appendix:partial}
\paragraph{Incomplete Source Dataset}
Table~\ref{tab:partial_sensitivity} contrasts performance on \textsc{Epilepsy} when using the entire source dataset (\(100\%\)) versus a much smaller portion (\(1\%\)). 
Notably, even \(1\%\) of \textsc{SleepEEG} or \textsc{ECG} is larger than the full training set of \textsc{Epilepsy}, indicating that a relatively small fraction of a homogeneous source domain can still provide meaningful pre-training benefits. 
However, the larger standard deviations in the \(1\%\) setting suggest that while small-scale pre-training can be beneficial, it may not always yield stable results—indicating a trade-off between data efficiency and robustness

\begin{table*}[!htb]
\small\centering\captionsetup{justification=centering, skip=5pt}
\caption{Comparative analysis of performance on \textsc{Epilepsy} with partial source dataset
\label{tab:partial_sensitivity}
}
\begin{tabular}{@{}clC{2.1cm}C{2.1cm}C{2.1cm}C{2.1cm}@{}}
\toprule
\textbf{Setting}            & \multicolumn{1}{c}{\textbf{Domain Adaptation}} & \textbf{Accuracy} & \textbf{Precision} & \textbf{Recall}  & \textbf{F1 score} \\ \midrule
\multirow{2}{*}{100\%}      &  \textsc{SleepEEG} $\rightarrow$ \textsc{Epilepsy}                                 & 0.958\std{0.002}  & 0.936\std{0.004}   & 0.935\std{0.004} & 0.931\std{0.003}  \\
                               &  \textsc{ECG} $\rightarrow$ \textsc{Epilepsy}                                      & 0.956\std{0.002}  & 0.934\std{0.004}   & 0.940\std{0.008} & 0.927\std{0.003}  \\ \midrule
\multirow{2}{*}{1\%} &  \textsc{SleepEEG} $\rightarrow$ \textsc{Epilepsy}                                 & 0.954\std{0.002}  & 0.923\std{0.006}   & 0.945\std{0.004} & 0.928\std{0.002}  \\
                               &  \textsc{ECG} $\rightarrow$ \textsc{Epilepsy}                                      & 0.943\std{0.009}  & 0.908\std{0.019}   & 0.937\std{0.004} & 0.913\std{0.012}  \\ \bottomrule
\end{tabular}
\end{table*}

\paragraph{Catastrophic Forgetting and Overfitting}
While adapting a model from one domain to another, two primary risks arise: \emph{catastrophic forgetting} of previously learned representations and \emph{overfitting} to the new domain’s limited data. 
In our cross-domain adaptation scenario, we mitigate these issues through a balanced loss objective that preserves domain-invariant features (via contrastive learning) while fine-tuning on the target domain (via cross-entropy loss). 
This joint learning scheme discourages drastic parameter updates that could wipe out essential knowledge from the source, and it helps prevent overfitting by maintaining a more generalizable feature space.

\paragraph{Cross-Domain vs.\ Multi-Source Pre-Training}
One possible approach to domain adaptation in time series is to merge data from multiple sources into a single, large pre-training set. 
Our one-to-one and one-to-many setups instead assume a relatively homogeneous source domain, making domain-invariant pattern learning more tractable and interpretable. However, when data originate from vastly different domains (e.g., EEG signals vs.\ industrial sensor logs), a many-to-one strategy may result in excessive heterogeneity that obscures useful temporal patterns. 
Developing robust solutions for large-scale, highly heterogeneous time series remains an open research direction, and we hope our systematic cross-domain experiments will serve as a foundation for future work aiming to unify multi-sourced data.

\subsection{Benchmarking Across Multivariate Time Series Classification Tasks}\label{appendix:mtsc}

We further evaluated our framework on ten multivariate time series classification (MTSC) datasets from the University of East Anglia (UEA) repository\footnote{\url{https://www.timeseriesclassification.com/}}\citep{bagnall2018uea}. 
For comparison, we use the reported performances of TST~\citep{zerveas2021transformer}, TS-TCC~\citep{eldele2021time}, TNC~\citep{tonekaboni2021unsupervised}, T-Loss~\citep{franceschi2019unsupervised}, TS2Vec~\citep{yue2022ts2vec}, MICOS~\citep{hao2023micos}, and Mgformer~\citep{wen2024mgformer} as presented in \citet{wen2024mgformer}.
We train our own model purely in a self-supervised manner, without using any additional external data (such as \textsc{SleepEEG} or \textsc{ECG}) or hyperparameter tuning process.

Table~\ref{tab:mtsc} highlights our approach’s performance on ten public multivariate time series datasets. The columns list classification accuracy for each baseline method and our proposed framework. Notably, the proposed method achieves competitive performance on a majority of the benchmarks. 
Table~\ref{tab:mtsc_ablation} examines the contribution of each architectural and training component in our proposed framework. By comparing these settings, we observe how multi-view fusion and the combined loss objective each drive consistent improvements across datasets, further validating the holistic design of our approach. Observing consistent gains when all three are combined supports our claim that integrating multiple feature views captures richer representations and yields stronger results across a wide range of time series applications.

\clearpage

\begin{table*}[!htb]
\small\centering\captionsetup{justification=centering, skip=5pt}
\caption{Comparative analysis of performance on MTSC
\label{tab:mtsc}
}
\begin{tabular}{@{}lcccccccc@{}}
\toprule
\multicolumn{1}{c}{\textbf{Dataset \textbackslash Method}} & \textbf{TST} & \textbf{TS-TCC} & \textbf{TNC} & \textbf{T-Loss} & \textbf{TS2Vec} & \textbf{MICOS} & \textbf{Mgformer} & \textbf{Proposed} \\ \midrule
\textsc{AtrialFibrillation}                                & 0.067        & 0.267           & 0.133        & 0.133           & 0.200           & 0.333          & 0.400             & 0.427\std{0.077}  \\
\textsc{BasicMotions}                                      & 0.975        & 1.000           & 0.975        & 1.000           & 0.975           & 1.000          & 1.000             & 0.995\std{0.012}  \\
\textsc{FingerMovements}                                   & 0.560        & 0.460           & 0.470        & 0.580           & 0.480           & 0.570          & 0.620             & 0.574\std{0.027}  \\
\textsc{Heartbeat}                                         & 0.746        & 0.751           & 0.746        & 0.741           & 0.683           & 0.766          & 0.790             & 0.755\std{0.005}  \\
\textsc{MotorImagery}                                      & 0.500        & 0.610           & 0.500        & 0.580           & 0.510           & 0.500          & 0.530             & 0.576\std{0.010}  \\
\textsc{RacketSports}                                      & 0.809        & 0.816           & 0.776        & 0.855           & 0.855           & 0.941          & 0.908             & 0.821\std{0.019}  \\
\textsc{SelfRegulationSCP1}                                & 0.754        & 0.823           & 0.799        & 0.843           & 0.812           & 0.799          & 0.890             & 0.886\std{0.010}  \\
\textsc{SelfRegulationSCP2}                                & 0.550        & 0.533           & 0.550        & 0.539           & 0.578           & 0.578          & 0.533             & 0.594\std{0.017}  \\
\textsc{StandWalkJump}                                     & 0.267        & 0.333           & 0.400        & 0.333           & 0.467           & 0.533          & 0.600             & 0.493\std{0.077}  \\
\textsc{UWaveGestureLibrary}                               & 0.575        & 0.753           & 0.759        & 0.875           & 0.906           & 0.891          & 0.893             & 0.827\std{0.035}  \\ \bottomrule
\end{tabular}
\end{table*}

\begin{table*}[!htb]
\small\centering\captionsetup{justification=centering, skip=5pt}
\caption{Comprehensive analysis of the proposed framework's components on MTSC
\label{tab:mtsc_ablation}
}
\begin{tabular}{@{}lcccc@{}}
\toprule
\multirow{2}{*}{\textbf{Dataset\hspace{1.0em}\textbackslash\hspace{1.0em}Model Component}}              & \multicolumn{2}{c}{Proposed method}                              & \multicolumn{2}{c}{w/o feature fusion}                                \\
                               & $\mathcal{L}_{CL}+\mathcal{L}_{CE}$ & use $\mathcal{L}_{CE}$ only & $\mathcal{L}_{CL}+\mathcal{L}_{CE}$ & use $\mathcal{L}_{CE}$ only \\ \midrule
\textsc{AtrialFibrillation}    & 0.427\std{0.077} & 0.413\std{0.088}               & 0.373\std{0.102}   & 0.360\std{0.113}               \\
\textsc{BasicMotions}          & 0.995\std{0.012} & 0.995\std{0.012}               & 0.995\std{0.012}   & 0.995\std{0.012}               \\
\textsc{FingerMovements}       & 0.574\std{0.027} & 0.564\std{0.018}               & 0.566\std{0.020}   & 0.564\std{0.018}               \\
\textsc{Heartbeat}             & 0.755\std{0.005} & 0.753\std{0.004}               & 0.745\std{0.006}   & 0.745\std{0.006}               \\
\textsc{MotorImagery}          & 0.576\std{0.010} & 0.572\std{0.016}               & 0.562\std{0.017}   & 0.548\std{0.029}               \\
\textsc{RacketSports}          & 0.821\std{0.019} & 0.817\std{0.017}               & 0.816\std{0.019}   & 0.816\std{0.019}               \\
\textsc{SelfRegulationSCP1}    & 0.886\std{0.010} & 0.878\std{0.021}               & 0.876\std{0.010}   & 0.868\std{0.019}               \\
\textsc{SelfRegulationSCP2}    & 0.594\std{0.017} & 0.576\std{0.015}               & 0.586\std{0.019}   & 0.568\std{0.021}               \\
\textsc{StandWalkJump}         & 0.493\std{0.077} & 0.440\std{0.061}               & 0.427\std{0.077}   & 0.427\std{0.077}               \\
\textsc{UWaveGestureLibrary}   & 0.827\std{0.035} & 0.823\std{0.033}               & 0.821\std{0.038}   & 0.812\std{0.038}               \\ \bottomrule
\end{tabular}
\\ \bigskip\bigskip
\begin{tabular}{@{}lcccccc@{}}
\toprule
\textbf{Dataset\hspace{1.0em}\textbackslash\hspace{1.0em}Model Component} & $\left\{\bm{h}_t,   \bm{h}_d\right\}$ & $\left\{\bm{h}_t,   \bm{h}_f\right\}$ & $\left\{\bm{h}_d,   \bm{h}_f\right\}$ & $\bm{h}_t$   only & $\bm{h}_d$   only & $\bm{h}_f$   only \\ \midrule
\textsc{AtrialFibrillation}    & 0.293\std{0.038}                      & 0.227\std{0.038}                      & 0.360\std{0.061}                      & 0.240\std{0.077}  & 0.413\std{0.057}  & 0.387\std{0.088}  \\
\textsc{BasicMotions}          & 0.970\std{0.012}                      & 0.990\std{0.015}                      & 0.995\std{0.012}                      & 0.965\std{0.030}  & 0.815\std{0.061}  & 0.990\std{0.015}  \\
\textsc{FingerMovements}       & 0.578\std{0.041}                      & 0.552\std{0.023}                      & 0.564\std{0.039}                      & 0.560\std{0.038}  & 0.566\std{0.010}  & 0.566\std{0.016}  \\
\textsc{Heartbeat}             & 0.722\std{0.013}                      & 0.726\std{0.011}                      & 0.733\std{0.007}                      & 0.738\std{0.016}  & 0.719\std{0.017}  & 0.731\std{0.013}  \\
\textsc{MotorImagery}          & 0.556\std{0.010}                      & 0.564\std{0.031}                      & 0.560\std{0.026}                      & 0.540\std{0.017}  & 0.500\std{0.001}  & 0.580\std{0.036}  \\
\textsc{RacketSports}          & 0.793\std{0.009}                      & 0.807\std{0.025}                      & 0.680\std{0.046}                      & 0.803\std{0.013}  & 0.676\std{0.020}  & 0.593\std{0.037}  \\
\textsc{SelfRegulationSCP1}    & 0.872\std{0.033}                      & 0.872\std{0.021}                      & 0.813\std{0.017}                      & 0.878\std{0.010}  & 0.509\std{0.018}  & 0.810\std{0.013}  \\
\textsc{SelfRegulationSCP2}    & 0.552\std{0.032}                      & 0.570\std{0.027}                      & 0.598\std{0.017}                      & 0.547\std{0.024}  & 0.517\std{0.038}  & 0.601\std{0.008}  \\
\textsc{StandWalkJump}         & 0.413\std{0.088}                      & 0.413\std{0.074}                      & 0.400\std{0.134}                      & 0.373\std{0.061}  & 0.413\std{0.088}  & 0.320\std{0.031}  \\
\textsc{UWaveGestureLibrary}   & 0.813\std{0.036}                      & 0.656\std{0.196}                      & 0.764\std{0.038}                      & 0.529\std{0.346}  & 0.752\std{0.027}  & 0.333\std{0.107}  \\ \bottomrule
\end{tabular}
\end{table*}

\end{document}